\documentclass[lettersize,journal]{IEEEtran}
\usepackage{amsmath,amsfonts}
\usepackage{algorithmic}
\usepackage{algorithm}
\usepackage{array}
\usepackage[caption=false,font=normalsize,labelfont=sf,textfont=sf]{subfig}
\usepackage{textcomp}
\usepackage{stfloats}
\usepackage{url}
\usepackage{verbatim}
\usepackage{graphicx}
\usepackage{cite}
\usepackage{tikz}
\usetikzlibrary{bayesnet}
\usetikzlibrary{arrows}
\usepackage{bm}
\begin{document}

\title{Ultra Low-Power, Lightweight, Probabilistic RSS-Based Path Reconstruction: A System for Landscape-Scale Bee Tracking}

\author{Christopher J. Noroozi, Joseph L. Woodgate, Michael Mangan and Michael T. Smith
\thanks{This work was supported by the Engineering and Physical Sciences Research Council [Project number: 2940233], the Eva Crane Trust [Grant reference: ECT\textunderscore202503432A], the University of Sheffield and the Grantham Center.
Christopher J. Noroozi, Michael Mangan and Michael T. Smith are with the School of Computer Science, University of Sheffield, Sheffield, UK (e-mail: cjnoroozi1@sheffield.ac.uk;  m.mangan@sheffield.ac.uk; m.t.smith@sheffield.ac.uk).
Joseph L. Woodgate was with the School of Computer Science, University of Sheffield, Sheffield, UK (e-mail: josephlukewoodgate@gmail.com).}%
}


\maketitle

\begin{abstract}
Applications in fields such as movement ecology, Internet of Things or robotics share the need for systems that localize devices that are too small and power constrained to implement GNSS (Global Navigation Satellite Systems). Alternative low-power localization methods often rely on only measurements of RSS (Received Signal Strength) to infer the AoA (Angle of Arrival) of a transmitted radio frequency signal, but are limited by range and the power demand of the large number of RSS measurements required to infer an accurate AoA. In this paper we address these issues with a novel RSS-based method for tracking ultra lightweight and low-power moving receivers across a complex landscape, achieved by using a minimal number of RSS measurements from simple rotating high-gain transmitters with a range of 300m, and applying probabilistic modelling to infer their AoA. The receiver's movement path is then modelled using a Gaussian process and reconstructed using doubly stochastic variational inference, resulting in $\sim$15m accuracy tracking of receivers weighing 38mg (including power source) over a scalable landscape range while consuming $<$180$\mu$W, increased to $\sim$10m accuracy at $<$600$\mu$W by taking more RSS measurements. We anticipate that this method will support fields such as the behavioural study of flying insect species, which we demonstrate by applying the system to track \textit{Bombus terrestris} nest return flights.
\end{abstract}


\section{Introduction} \label{chapter:introduction}
\IEEEPARstart{I}{n} several fields, the ability to continuously measure the position of very lightweight devices over large areas is of great importance. In movement ecology, automated tracking of an animal's position over time has been key in advancing knowledge of fundamental aspects of animal behaviour \cite{kays2015terrestrial} and informing conservation efforts \cite{katzner2020evaluating}. Tagged tracking methods are often required for landscape-scale tracking, and also show promise in the implementation of biologging sensors to infer behaviour along with position \cite{williams2020optimisingbiologgers}. To enable the study of very small animals such as insects, these tags must be extremely lightweight and low-power. The ubiquity of personal mobile devices has driven widespread research in the use of low-energy, lightweight Internet of Things (IoT) sensor networks \cite{alahi2023integrationofIoT} requiring device localization to deliver services to users, including navigation and medical services \cite{esmaeili2024improvingwifilocalization}. Emerging fields such as smart dust or swarm robotics also demand novel localization of very small devices \cite{romer2003lighthouse, chen2022survey}. All of these applications share a common instrumentation problem: measuring position continuously and accurately under constraints of minimal mass and power consumption.

The tracking of such devices is most commonly performed using GNSS (Global Navigation Satellite Systems). This however requires substantial hardware and battery capacity. Grenier \textit{et al.} \cite{grenier2023asurgveyonGNSS} concludes that GNSS is often the most power consuming sensor in embedded systems, with the smallest GNSS units commercially available such as the U-Blox UBX-M10150-CC \cite{ublox_m10150cc_web} requiring 8mW to function. In highly hardware constrained tracking applications where minimal power consumption is key, the high power demand and added weight of a sufficient enough battery for GNSS to function precludes tracking of smaller subjects such as the majority of flying insect species \cite{wilson2021animallifestyle}. Many applications also require tracking in situations where GNSS is not available such as indoor tracking of mobile devices \cite{tiglao2021smartphone}, or underground localization of UAVs which faces both an absence of GNSS availability and strict power constraints \cite{zhang2022demand}.

Existing methods for automated, landscape-scale tracking of flying insects have relied on either Very High Frequency transceivers \cite{wikelski2010large} or harmonic radar \cite{riley1996tracking, woodgate2016life}. The former requires a relatively heavy tag ($>$100mg), causing energetic costs on the animal \cite{hagen2011space} and precluding the tracking of widely studied species such as \textit{Bombus terrestris} which Stelzer \textit{et al.} \cite{stelzer2010winter} show carry up to 70mg of weight per foraging bout. The latter is prohibitively expensive and struggles in complex environments. Dressler \textit{et al.} \cite{dressler2019ultralowpower} note that since most uses of sensor technology for wildlife monitoring are battery powered, energy consumption is a key consideration. A system that is suitable for tracking such subjects therefore must consume minimal power and use tags weighing $<70mg$. 

Received signal strength (RSS) based localization methods are an ideal alternative for miniaturised systems due to the minimal signal processing and hardware overhead required to take RSS measurements \cite{nowak2016fundementallimits, coluccia2010ml}. Distance-free RSS methods exploit the anisotropy of a directional antenna's radiation pattern (ARP) to infer the angle of arrival (AoA) of a transmitted signal. Several bearing observations from spatially separated transceivers can then be used to triangulate a device or reconstruct the path it took. However, as reviewed in Section \ref{chapter:relatedwork}, existing RSS-based methods either require a prohibitively large number of RSS measurements per inferred AoA - incurring energy costs incompatible with ultra low-power receivers - or are limited in range, preventing deployment at landscape scale.

To address these measurement limitations, we present a novel RSS-based, distance-free method of tracking a moving, low-power, lightweight receiver across a complex, unknown landscape. By using a sparse duty cycle to receive short bursts of Bluetooth Low-Energy (BLE) packets from rotating high-gain transmitters placed across the landscape, the receiver requires far less power than previous RSS-based AoA methods. Rather than seeking the peak or null of the ARP over a full rotation, which requires dense RSS measurement, we instead present and evaluate three methods to probabilistically model and infer the AoA of the received packets based on their change in RSS over time. We then perform a form of `probabilistic triangulation' from these non-concurrent, uncertain AoA predictions to reconstruct the receiver's full movement path. We do this by modelling the receiver's path using a Gaussian process (GP), and perform inference via doubly stochastic variational inference, providing an expression of the path's uncertainty over time.

The result is a 38mg (including power source) archival receiver, whose path at a landscape scale can be reconstructed using packets received from simple rotating transmitters each with a range of 300m, with a positional accuracy of $\sim$15m mean absolute error while consuming $<$180$\mu$W, improving to $\sim$10m at $<$600$\mu$W.

We validate the ability of the system to track the receiver by applying it in several ground truthed localization tasks, evaluating AoA \& path reconstruction accuracy and the effect of varying the number of RSS measurements per AoA inferred. We also demonstrate a use case of the system in movement ecology by deploying it in a field trial to track the return flight of a foraging \textit{Bombus terrestris} (buff-tailed bumblebee). 

This paper is organized as follows. In Section II the state of the art in the field of RSS-based localization is explored. In Section III the novel hardware for our proposed ultra low-power, lightweight tracking system is described. Section IV presents three approaches for probabilistically inferring an AoA based on received RSS measurements, and Section V describes the path reconstruction algorithm that uses these AoAs to infer the receiver's path. Section VI outlines the experimental set-up for validating the performance of the system, the results of which are presented in Section VII. In Section VIII the conclusions are drawn.

\section{Related Work} \label{chapter:relatedwork}
To explore alternatives to GNSS, Hayward \textit{et al.} \cite{hayward2022indoorlocationtech} define four groups of localization methods: acoustic, inertial, visual and radio frequency (RF) based (although outside this taxonomy, other methods do exist such as using magnetic fields \cite{sheinker2018localization}). Of these, RF based approaches are the most commonly used.

Localization via RF has been achieved using: the phase difference between signals received by multiple antenna to obtain the Angle of Arrival (AoA) of a signal \cite{9093137}, the difference in time of arrival (ToA) between multiple received signals \cite{zouToA}, measurements of received signal strength (RSS) or a hybrid of these methods \cite{POURKABIRIAN2023103177, liuToAAoARSS, nyantakyi2024acga}. ToA requires precise timing \& synchronization, and accessing the phase difference of received signals requires additional \& complex hardware beyond a single antenna \cite{coluccia2010ml}. Instead, purely RSS based localization is ideal for miniaturized low-power systems due to the relatively low complexity of signal processing and calibration required \cite{nowak2016fundementallimits}. 

Distance-based methods of localization using RSS attempt to infer the relation between RSS and the transmitter-receiver distance using a path loss model \cite{bandiera2015cognitive}, and therefore find the distance from several transceivers and trilaterate the position of a device \cite{luomala2022adaptiverangebased}. However a key challenge of using RSS measurements to infer distance is the effect of other factors such as antenna orientation or the significant noise induced by cluttered environments and multipath fading \cite{zanella2016bestpractice}. These severely impact the reliability of distance estimates. Fingerprinting techniques have improved distance based localization by first learning the RSS at a number of known locations from several static beacons \cite{kriz2016improving}. RSS measurements at the device to be localized, either a single vector of measurements or a series of measurements over time \cite{shin2022received}, are then compared with these ``fingerprints'' via clustering \cite{yiu2017wirelessrssifingerprinting}, deep learning \cite{subedi2017practical}, filtering \cite{peng2021new} or other methods to infer the device's location. This method has resulted in high-accuracy localization, but learning the required fingerprints leads to a poor ability to generalize and scale for unknown and changing environments without laborious recalibration, which much research aims to streamline \cite{nagy2020rssdirectionallocalization}.

Distance-free methods of RSS-based localization exploit the anisotropy of a non-isotropic antenna's RF radiation pattern (ARP) coupled with a device's movement to infer the relative AoA, sometimes also known as Direction of Arrival, of a transmitted signal \cite{varotto2021transmitter}. Multiple AoAs, essentially bearing observations, from different transceivers can then be used to either triangulate a device \cite{honma2018RSSbasedDOAestimation} or reconstruct a full trajectory via filtering and smoothing frameworks, such as pseudo-linear or Kalman filter-based approaches \cite{rs16234536, rs13152915, qian2024maneuvering}.

These methods often seek the null \cite{tamura2023nullsteering} or peak \cite{hood2010estimating} of an ARP facing a known angle to infer the AoA. As RSS varies with AoA and distance, this requires very frequent measuring at the device to capture enough of the ARP to find the RSS minima or maxima respectively, incurring significant energy cost that is not feasible for a low-power device.

To infer AoA with fewer RSS samples, other methods have used a form of monopulse in which RSS measurements are taken from multiple antenna with differently oriented ARPs at the same location \cite{jiang2013ALRD}. The difference between measurements of RSS from these antenna is then compared with the known difference in gain across their ARPs to infer a relative AoA \cite{malajner2015aoamultiplemono}. Many such methods \cite{jais2015review} have seen success in low-power, lightweight device localization applications, especially in animal tracking, such as the triangulation of bats in the wild \cite{duda2018bats}. The system was able to perform high resolution tracking in a relatively small area of a few hectares. Extensions to such approaches have lead to inference of distance alongside AoA \cite{li2019cost}, effectively allowing for localisation from a single node consisting of only three antenna, or improvements in AoA accuracy to within 1$^\circ$ \cite{zheng2021rssi}. Common across all these methods is low gain of the antenna used, leading to a poor ability to scale across large areas due to the prohibitively large number of antennas needed to cover a large portion of the landscape and the reduced angular precision from their wide half-power beamwidth (HPBW). The use of high-gain yagi antenna have resulted in greater transmitter range in a study tracking monarch butterflies \cite{fisher2021locating}. However, whilst high-gain antennas extend range, their narrow HPBW means that achieving full 360$^\circ$ coverage from a single node would require a prohibitively large number of antenna elements. Fisher \textit{et al.} \cite{fisher2021locating} mentions that 9 receiving stations - each using 4 antennas resulting in 36 antennas total - would be required to cover an area of 500m × 500m using their method, limiting the practicality of landscape-scale deployment.

To overcome this Varotto \textit{et al.} \cite{varotto2021activesensingsearchtracking} identifies the need for motorized directional antennas, used in fingerprinting approaches \cite{burton2022orientation} and robotics tracking applications in which the orientation of the ARP indicates the AoA the tracked target lies upon. These methods once again rely on capturing a large number of RSS measurements to find the peak RSS from an ARP's main lobe \cite{ghosh2019arrest} (200 RSS measurements per antenna rotation) or reconstruct a portion of the antenna's ARP \cite{graefenstein2009wireless} (RSS measurement every 11$^\circ$-16$^\circ$ of antenna rotation).

All of the above RSS based tracking methods either perform localization using too many RSS measurements to be feasible for an ultra low-power receiver, or struggle to scale across a larger landscape scale due to low-range or cumbersome antenna.

The system we present in this paper addresses these issues by performing AoA inference with a burst of as few as 3 RSS measurements, received from simple rotating high-gain antenna with a range of 300m, sufficient to cover a landscape scale. This is accomplished by probabilistically modelling the change in measured RSS across the burst. We then model the receiver's movement path from several of these AoA predictions using a Gaussian process, and perform inference using doubly stochastic variational inference, resulting in full path reconstruction.

\section{Hardware} \label{chapter:hardware}
Our system reconstructs the movement path of a 38mg archival receiver mounted on the subject that we want to track. The receiver logs packets and their RSS from any number of rotating transmitters placed in known, stationary positions across the landscape, which are later used in Section \ref{chapter:pathreconstruction} for AoA inference.

\begin{figure}[!t]
    \centering
    \includegraphics[width=3in]{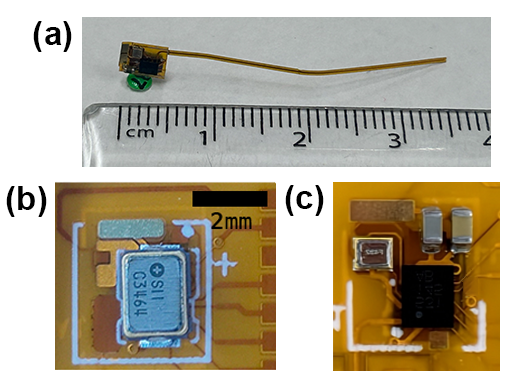}
    \caption{The 38mg supercapacitor powered receiver that the system will track. (a) The receiver, affixed to a green queen bee marking tag. Ruler provided for scale. (b) One side of the receiver showing the supercapacitor and a pad for charging, the white outline bounds the receiver on the PCB, the tracks and pads outside this region are used for programming the DA14531 and then removed. (c) Other side of the receiver, showing the DA14531, crystal oscillator, multi-layer ceramic capacitors, ground pad for charging and a pad to trigger an interrupt for reading data off the receiver.}
    \label{fig:receiver}
\end{figure}

The receiver shown in Fig.~\ref{fig:receiver} was designed to be as low profile and lightweight as possible. It consists of a flexible polyimide PCB upon which the DA14531 System on Chip \cite{da14531}, one of the smallest and lowest power commercially available 2.4GHz BLE enabled chips, is surface mounted along with a 32MHz crystal oscillator and two 22$\mu$F multi-layered ceramic capacitors (a buffer for the high demand receiving duty cycle). The device uses a 3cm long, 0.2mm width copper PCB trace as a monopole antenna. For the power source we use a 24mg, 11mF Seiko CPH3225A supercapacitor \cite{CPH3225A} which results in a 16mm$^{2}$ (excluding antenna surface area) package weighing only 38mg. Other power sources are feasible such as a button battery, thin film solar cell or microbatteries.

The receiver scans for and stores 2.4GHz BLE packets sent from the transmitters along with their RSSI (Received Signal Strength Indicator), using a short and therefore low-power 7ms receiving period (as this is the minimum window of time needed to receive the majority of packets) to capture a packet from all transmitters in range and then enter a hibernation mode until the duty cycle repeats. The RSSI measured by the DA14531 is a chipset-specific indicator proportional to the RSS of the packets. At a receiving power expenditure of 5mA, the 11mF supercapacitor powered receiver can perform over 500 receiving duty cycles over its powered lifespan. Bursts of these receiving cycles are used to sample sets of $k$ packets which are used for AoA inference once the receiver has been retrieved and the data read from its storage.

The total duration of the burst and the intervals within this duration that the receiver turns on can be varied to change $k$, and therefore balance power consumption \& device lifespan with AoA inference accuracy and resulting path reconstruction accuracy. We investigate the effects of varying these parameters through the experiments described in Section \ref{chapter:experimentaldesign}.

The receiver is charged prior to tracking using a pair of terminals on its corner. The data the receiver has stored can be read using an interrupt pad to change it to a transmitting mode, meaning the receiver is archival.

\begin{figure}[!t]
    \centering
    \includegraphics[width=3in]{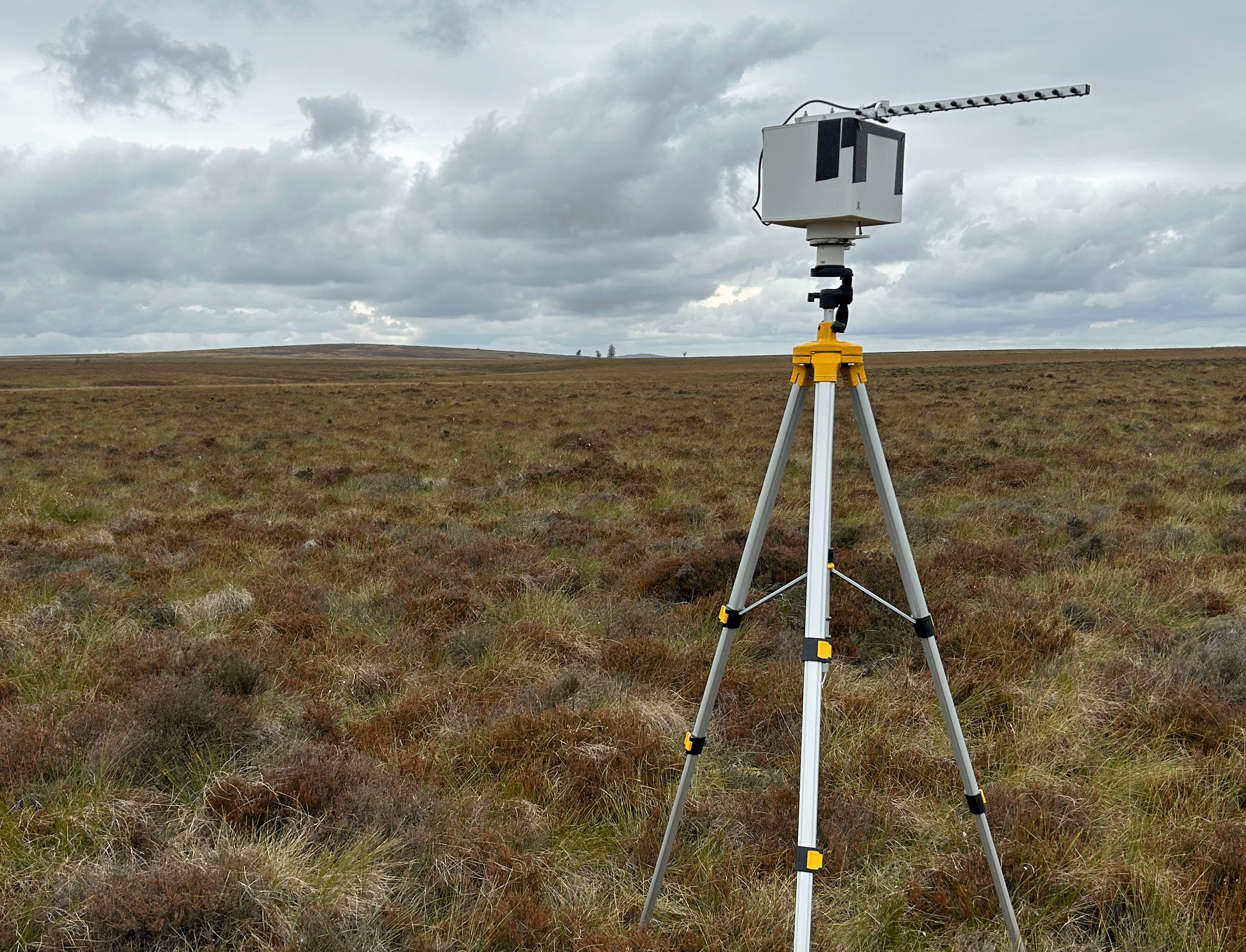}
    \caption{Our simple 30rpm rotating high-gain transmitter with a 14dBi Yagi Uda antenna mounted on a 1.5 meter tall tripod in the field, transmitting packets used to track the receiver.}
    \label{fig:transmitter}
\end{figure}

The rotating transmitters shown in Fig.~\ref{fig:transmitter}, mounted on tripods at known locations across the landscape, use a 14 dBi high-gain Yagi-Uda antenna to broadcast BLE directed advertising packets to localize the receiver. This type of advertising packet contains only the source (transmitter) and destination (receiver) Bluetooth device addresses, and was chosen as it is specifically addressed to the receiver, allowing for the receiver to ignore all other BLE traffic, reducing the power it consumes when processing received packets. Further, directed advertising packets can be updated quickly and have a high chance of being received at any given receiving duty cycle due to their short, aggressive advertising interval. Directed advertising packets have no data payload; to allow for transmission of data to the receiver, we encode within the 6 byte BLE advertising address the angle $\bm{\gamma}$ of the transmitter's rotation (relative to North), the time of transmission and the id of the transmitter (from which the known location of the transmitter can be found). The packets are constantly transmitted, and updated every 10ms (approximately every 1.8$^\circ$ of rotation at 30rpm). These packets can be received by our BLE receiver up to 300m away from the transmitter, and ideally the receiver should be within range of at least two transmitters for localization.

The transmitters, not including the tripod they are mounted on, weigh 3 kg and measure 20 cm × 20 cm × 20 cm, with the Yagi antenna extending an additional 35 cm forward. A 30 rpm Pololu 150:1 Metal Gearmotor \cite{pololu_2828} drives the transmitter's rotation while a Fanstel BT832XE Bluetooth Module \cite{fanstel_bt832xe} reads the motor's encoder for the angle of rotation and constructs the BLE packets transmitted. A magnet is placed due North on the tripod, such that a Hall effect sensor on the rotating portion of the transmitter can align the angle of rotation with North. A 12V 8Ah LiFePo4 within the transmitter can power it for over 10 hours. The simple design of the transmitters allows several to be easily deployed across a landscape, scaling the trackable area and providing overlapping coverage in the presence of clutter.

\section{Angle of Arrival Inference} \label{chapter:angleinference}
\begin{figure}[!t] 
    \centering
    \includegraphics[width=3.5in]{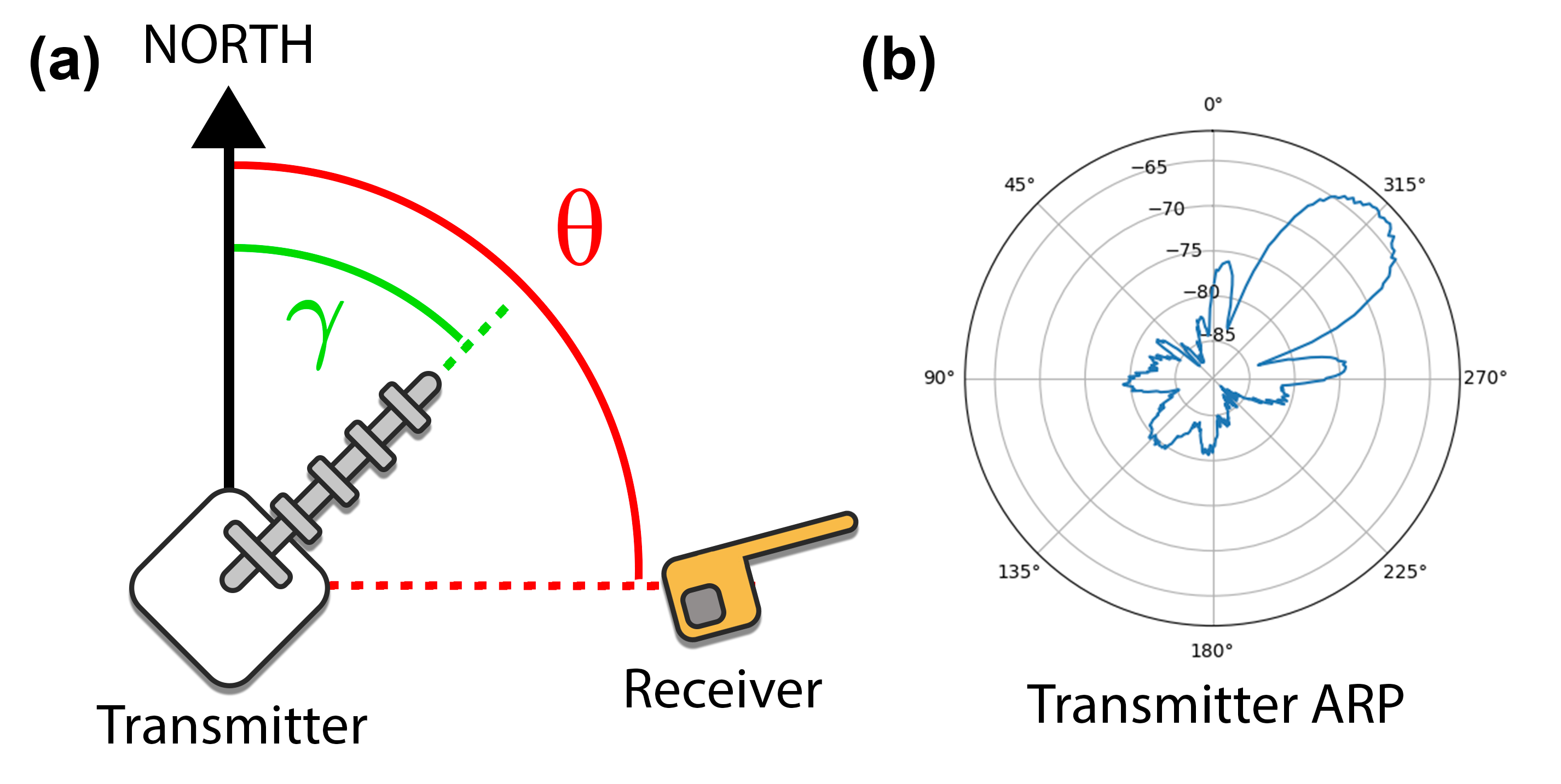}
    \caption{The problem of inferring receiver angle from packets received. (a) Definitions of angles. We seek $\theta$, the angle the receiver lies upon from the transmitter relative to North. $\gamma$ is the angle relative to North that the transmitter and therefore its ARP is oriented towards, this is always known as it is encoded in the packets received. (b) Polar plot of the ARP of our transmitter's 14 dBi Yagi Uda antenna oriented 45$^{\circ}$ from North, to show how the antenna's gain varies by angle. The ARP was measured empirically as described in Section \ref{chapter:trainingdata}, and averaged at each angle to remove noise as suggested by Kim \textit{et al.} \cite{kim2013smartphone}.}
    \label{fig:gammathetaexplained}
\end{figure}

We must infer the AoA of at least two transmitters from the receiver's location for use in path reconstruction, using the packets and their RSS received by the limited number of duty cycle bursts performed by the receiver. We wish to infer, as shown in Fig.~\ref{fig:gammathetaexplained}(a), the unknown angle $\theta$ relative to North that the receiver lies upon when receiving a burst of $k$ RSS measurements, $\bm{y}$, taken at known angles $\bm{\gamma}$ of the transmitter ARP's (shown in Fig.~\ref{fig:gammathetaexplained}(b)) rotation relative to North (encoded in the received packets) i.e. $p(\theta|\bm{y},\bm{\gamma})$. 

Ideally we should minimize the value of $k$ required to infer an accurate AoA, as this will minimize the power consumption of the receiver. We present here three different approaches for inferring an AoA from a limited number of RSS measurements, and evaluate their performance in field experiments described in Section \ref{chapter:AoAInferenceExperiment}.

\subsection{Full Probability Distribution Method} \label{chapter:fullprobmethod}
Similar to the approaches applied by Ghosh \textit{et al.} \cite{ghosh2019arrest} and Varotto \textit{et al.} \cite{varotto2021transmitter} - in which patterns in RSS measurements are matched with the peaks/nulls of a known high-gain antenna's ARP - we note that the anisotropy of our high-gain antenna generates unique changes in $\bm{y}$ across its rotation irrespective of distance, notably around its $\sim$40$^{\circ}$ H-plane HPBW. These changes in $\bm{y}$ will be offset by the known $\bm{\gamma}$ and unknown $\theta$. Therefore we can infer $\theta$ from observing $\bm{y}$ and using the known values of $\bm{\gamma}$. We model this using a Bayesian approach,
\begin{align}
    p(\theta|\bm{y},\bm{\gamma}) = \frac{p(\bm{y}|\theta,\bm{\gamma})p{(\theta})}{p(\bm{\bm{y}}|\bm{\gamma})}.
\end{align}

The true signal strengths $\bm{t}$, transmitted from the rotating high-gain antenna in the direction of the receiver, depend only on the direction of the receiver, $\theta$, offset by the known angles the transmitter is facing, $\bm{\gamma}$,
\begin{align}
\bm{t} = [t(\theta - \gamma_1), t(\theta - \gamma_2) ,\text{..., }t(\theta-\gamma_{k})]^{\top}.
\end{align}

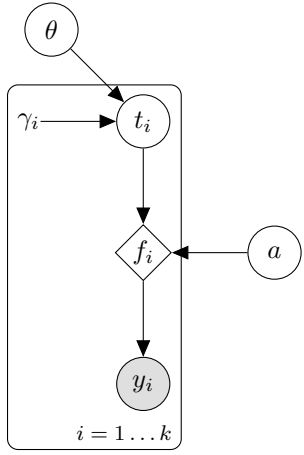
\begin{figure}[h]

\centering
\begin{tikzpicture}
 \node[obs] (y) {$y_i$};
 
 \node[det, above=of y] (f) {$f_i$};
\node[latent, above=of f] (tr) {$t_i$};  
 \node[const, left=of tr] (g) {$\gamma_i$}; 
 \node[latent, right=of f] (a) {$a$};
 \node[latent, above left=of tr] (t) {$\theta$};

 \edge {tr} {f} ;
 \edge {a} {f} ;
 \edge {f} {y} ; 
 \edge {t,g} {tr};%

 \plate {pl} {(y)(f)(g)(tr)} {$i=1\dots k$} ;
\end{tikzpicture}  
     \caption{The simple graphical model relating the angle $\theta$ the receiver lies upon relative to North and the observed RSS measurements $\bm{y}$ - a set of $k$ RSS observations taken at angles $\bm{\gamma}$ of transmitter rotation relative to North. We assume that the `true' unattenuated signal strength, $t_i$, depends on just the angle $\theta - \gamma_i$. We assume in this model that all the RSS measurements were taken close enough together in time that they all were generated by angles close to $\theta$ and experienced the same attenuation $a$ to result in signal strengths $\bm{f}$. Finally the RSS measurements actually observed by the receiver, $\bm{y}$, are modelled by assuming some Gaussian noise corrupts this attenuated signal.}
    \label{fig:RFpattern}
\label{fig:graphicalmodel}
\end{figure}

They are then affected by some unknown attenuation $a$, which is primarily due to transmitter-receiver distance, but can also result from obstructions or multipath interference. An important side note is that, we assume that the $k$ measurements of RSS within the burst are taken close enough together in time that they are affected by the same constant value of $a$. The most informative part of the pattern is around the narrow main lobe, and as the $40^{\circ}$ HPBW passes across the receiver in only 220ms, it is likely that the attenuation is almost constant. Finally the actual observations of RSS made by the receiver are assumed to be corrupted by some uncorrelated, zero-mean Gaussian noise with variance $\sigma^{2}$. This model is described with plate notation in Fig.~\ref{fig:graphicalmodel}.

To use this approach we must first empirically profile the ARP of our transmitter by taking several RSS measurements from a range of angles of rotation $\gamma \sim \text{Uniform}(0^\circ, 360^\circ)$ at a known $\theta = 0$. We can interpolate between the  average signal strengths at each sampled angle to allow us to estimate $t(\theta - \gamma)$ - i.e. what the transmitter's signal strength will be when the tag is at bearing $\theta$ from the transmitter, and the transmitter is pointing at angle $\gamma$. Note that these empirical transmitter RSS values are themselves attenuated, but because the receiver was stationary during this data collection, this is a constant offset; and, as we're only interested in the relative change in signal strength, doesn't affect inference.

So a single observation of the RSS, $y_i$, is modelled as,
\begin{align}
    p(y_{i}|\theta,\gamma_{i},a) = N(y_{i}|t(\theta-\gamma_{i})-a,\sigma^{2}).
\end{align}
We have several observations ($y_1 ... y_k$) for a given value of $\theta$. $a$ is unknown but, by putting an improper flat prior on $a$, we can integrate it out in closed form (see Supplementary Material SI for detailed derivation). Defining $d = ||(\bm{t}-\bar{\bm{t}})-(\bm{y}-\bar{\bm{y}})||_2$, we can write,
\begin{align}
    p(\bm{y}|\theta,\bm{\gamma}) = N(d|0,\sigma^{2}).
\end{align}

So to compute $p(\theta|\bm{y},\bm{\gamma}) = p(\bm{y}|\theta,\bm{\gamma})p(\theta)/p(\bm{y}|\bm{\gamma})$, we note that we have assumed a uniform prior on $\theta$, and that the denominator is constant w.r.t. $\theta$, which means we can compute $p(\bm{y}|\theta,\bm{\gamma})$ over a uniform grid and normalise.

To summarise the calculation: We compute for a uniform grid of angles from $\theta \sim Uniform(0^{\circ}, 360^{\circ})$, the predicted transmitter signal strength $t(\theta-\gamma_{i})$ associated with each observation's transmitter angle $\gamma_i$. We then mean centre $y_{i}$ and $t_{i}$ and compute the probability of our $k$ observations as,

\begin{align} \label{eq:thetaloglikelihood}
    p(\bm{y}|\theta,\bm{\gamma})=C \times \exp\Big[-\sum^{k}_{i=1}\frac{((t_i-\bar{t})-(y_{i}-\bar{y}))^{2}}{\sigma^{2}}\Big].
\end{align}

\begin{figure}[!t]
    \centering
    \includegraphics[width=3.5in]{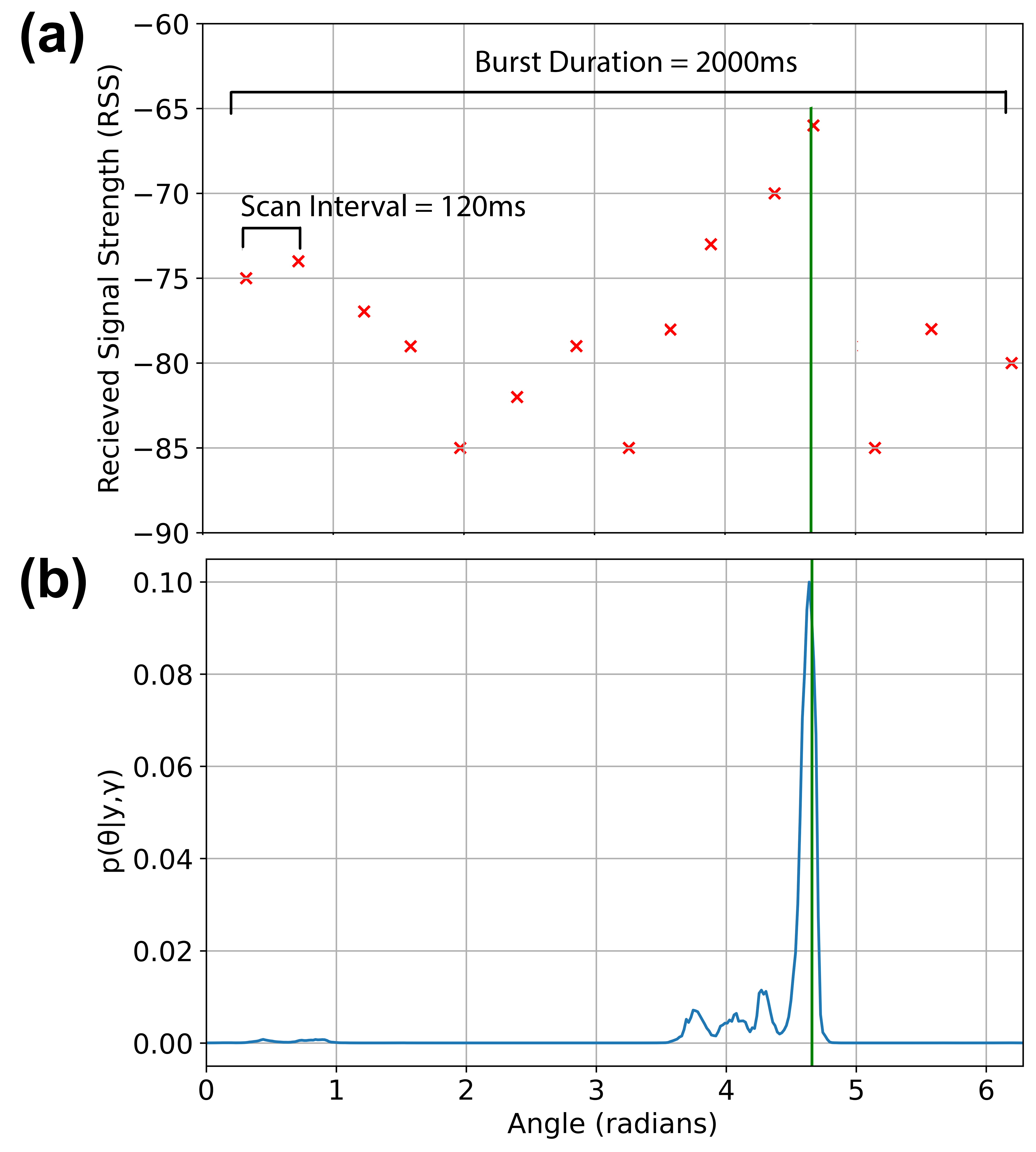}
    \caption{Example of AoA prediction returned by full probability distribution approach. (a) An example of a burst of RSS measurements (red), taken at a true angle of 4.6 radians (green), of packets received from a single rotating high gain transmitter. This burst was 2000ms in duration, and receiving duty cycles were performed at an interval of 120ms, resulting in a total of $k = 15$ samples taken. (b) The calculated probability distribution of the predicted angle $p(\theta|\bm{y},\bm{\gamma})$ from the full probability distribution method given the RSS measurements taken (blue). Note that the probability density is highest around the true angle, showing that our $p(\theta|\bm{y},\bm{\gamma})$ closely represents the true angle $\theta$ that generated the RSS observations $\bm{y}$ at angles of transmitter rotation $\bm{\gamma}$.}
    \label{fig:burstexample}
\end{figure}

Finally we normalise these probabilities, across $\theta$, to give the distribution of $p(\theta|\bm{y},\bm{\gamma})$ - our probability distribution of the angle the tag lies upon from the transmitter relative to North. An example of a calculated distribution of $p(\theta|\bm{y},\bm{\gamma})$ given an RSS measurement burst is shown in Fig.~\ref{fig:burstexample}.

We can use the full probability distribution of $p(\theta|\bm{y},\bm{\gamma})$ as an AoA observation for path reconstruction. This captures not only the AoA we predict the receiver lies upon from the transmitter, but also the uncertainty in our prediction, which is useful as even highly uncertain predictions may still contribute information to the path of movement the receiver took through the landscape.

\subsection{Rejection Sampling Method} \label{chapter:rejectionsampling}
$p(\theta|\bm{y},\bm{\gamma})$ can also be approximated using a training set of signal strengths to perform rejection sampling.

\begin{figure}[!t]
    \centering
    \includegraphics[width=3.5in]{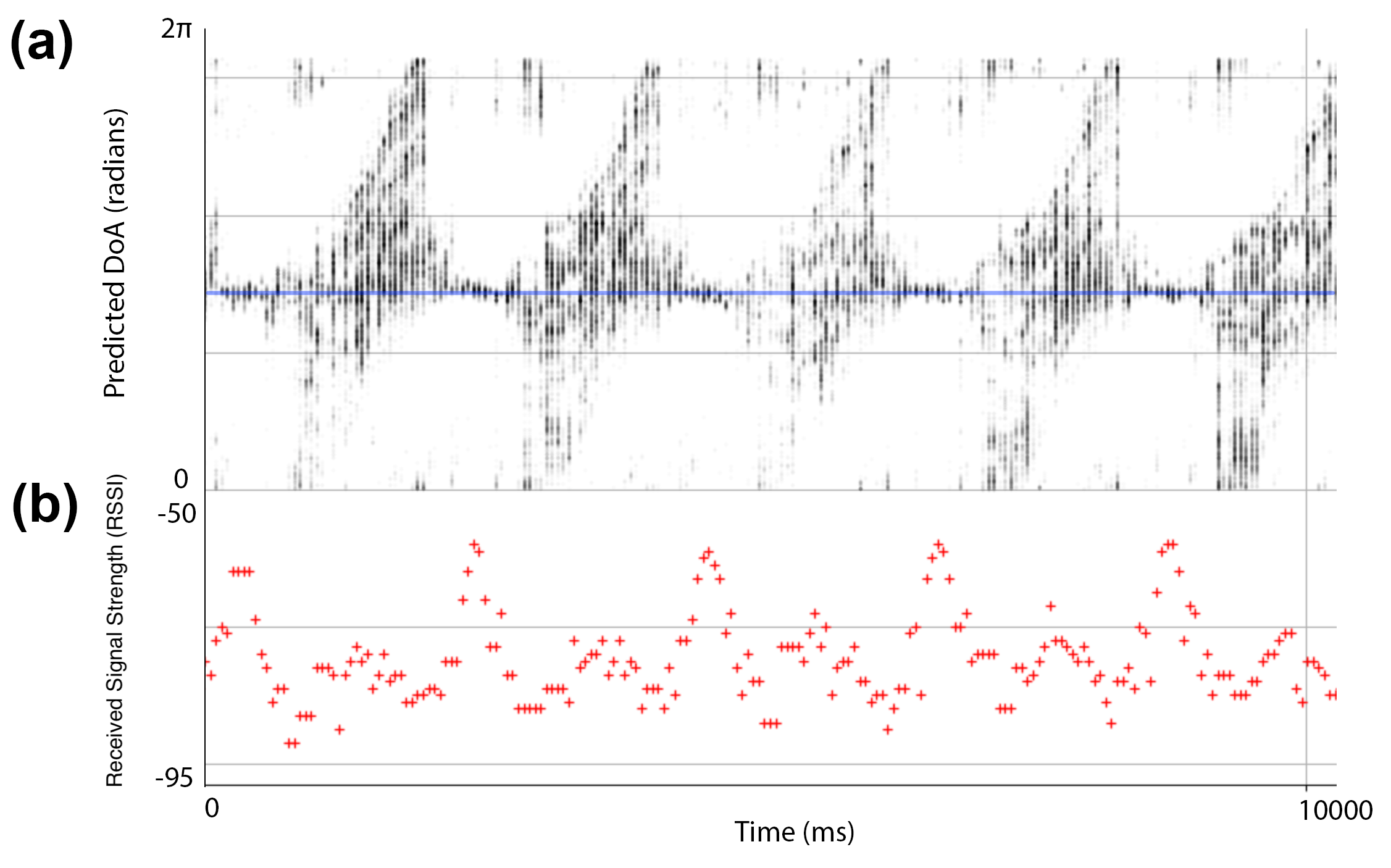}
    \caption{Example of AoA predictions returned by rejection sampling approach. (b) A plot of RSS measurements made when the receiver was positioned $\pi$ radians relative to North from the transmitter and captured packets (red) every 50ms for 10 seconds. (a) The data in (b) is subsampled, such that, at each time point just 5 packets spaced evenly 1 second to either side are used with the rejection sampling approach: The angles associated in the training set with similar relative signal strengths are plotted in black (approximating $p(\theta|\bm{y},\bm{\gamma})$). One can see that, from RSS measurements taken around the main lobe of the transmitter's antenna (where the RSS peaks), the samples from $p(\theta|\bm{y},\bm{\gamma})$ are confidently centred around the true angle $\theta = \pi$ (blue line). Samples taken away from the main lobe are instead highly uncertain and distributed across many values of $\theta$.}
    \label{fig:RejectionSamplingExample}
\end{figure}

To do this we generate a table $R$ in which each row is a set of RSS measurements (associated with a known $\theta$ and $\bm{\gamma}$) that were received at a set of angles whose relative differences match those in our observations $\bm{\gamma}$. We then select all the rows of $R$ in which the training set's mean-centred signal strengths are almost equal to our mean-centred signal strength observations. For these selected rows, their associated values of $\theta$ are taken and used as empirical sets of samples from $p(\theta|\bm{y},\bm{\gamma})$. These are plotted in Fig.~\ref{fig:RejectionSamplingExample}(a) in which we estimate $p(\theta|\bm{y},\bm{\gamma})$ at each observation time by assuming we only have access to five packets spaced uniformly at times [-1s, -0.5s, 0s, +0.5s, +1s] (enough time for the transmitter to rotate one radian on either side of the prediction time).

As demonstrated by Fig.~\ref{fig:RejectionSamplingExample}{a}, RSS measurements shown in Fig.~\ref{fig:RejectionSamplingExample}(b) taken around the HPBW of the antenna (where the RSS peaks) generate samples from $R$ that are densely centred around the true $\theta$, but away from this unique part of the ARP there are many likely values of $\theta$. This demonstrates a future direction for this work; using an active learning approach to target the most informative periods.

If we wish to avoid using the full distribution of $p(\theta|\bm{y},\bm{\gamma})$ for later path inference, and instead use a single estimate of $\theta$, we can select only confident predictions from multiple bursts by only including those with a small standard deviation, and use the circular mean of these as a single angle observation for later path reconstruction. We found rejection sampling highly sensitive to parameters such as the number of signal strengths per burst (therefore the size of each row in $R$) and the threshold of how similar our observations $\bm{y}$ must be to those in a row for its associated $\theta$ to be sampled.

\subsection{Peak RSS Method} \label{chapter:peakfinding}
A simple heuristic can also be used for AoA inference by noting that, if we were to receive a very large number of packets, $\theta = \gamma$ at the angle $\gamma$ that causes the transmitter to face the receiver and therefore produces the peak RSS across a full rotation of the transmitter. However, we do not run the receiving duty cycle often enough for one of the packets to be received exactly when $\theta = \gamma$.

To approximately find the this peak, we smooth $\bm{y}$ across its corresponding values of $\bm{\gamma}$ using a Savitzky–Golay filter, and use the peak of the smoothed curve to be our estimate of the AoA $\theta$. This method is simple to implement but highly susceptible to noise and the properties of the duty cycle burst we use to measure $\bm{y}$. A burst that is too short or does not have enough measurements $k$ to have a high probability of measuring a value of $y$ around the HPBW will not produce a good estimate of $\theta$.

\section{Path Reconstruction} \label{chapter:pathreconstruction}
\subsection{Choice of Model}
Triangulation from the AoA predictions over time inferred in Section \ref{chapter:angleinference} is not feasible as predictions from multiple transmitters are not necessarily concurrent, and the predictions made are often highly uncertain due to RSS noise. We wish to infer $p(F|Y)$, the path $F$ the receiver took over time that generated the AoA observations $Y$. This is a Bayesian inference problem,
\begin{align}
    p(F|Y) = \frac{p(Y|F)\;p(F)}{p(Y)}.
\end{align}

We model the receiver's path as a Gaussian process (GP) - a Bayesian model that can be used for regression, defined by a mean function and covariance/kernel function \(K(x,x')\). We place GP priors on the x \& y coordinates of the receiver as functions of time. Our kernels are constructed with the assumption that the axes are independent \textit{a priori}; however we model these with variational inference using a Gaussian endowed with a full rank covariance matrix, such that in the posterior the two axes can interact. A GP is a useful choice of model as it allows \textit{a priori} knowledge of the tracked subject's plausible movement to be included via the choice of \(K\) and provides a convenient expression of uncertainty \cite{rasmussen1997evaluation}.

We identify two suitable kernel functions, parametrized by length scale $l$ and scale factor $s^{2}$. The first is the commonly used EQ kernel,
\begin{align}\label{eq:ExponentiatedQuadratic}
K_{EQ}(\mathbf{x}, \mathbf{x}') = s^{2} \text{ exp}\Big(-\frac{|\mathbf{x}-\mathbf{x}'|^2}{2l^2}\Big).
\end{align}

The second is the integral of the EQ kernel, which places the EQ prior on the velocity. An important difference is that while the posterior of the EQ kernel will tend towards the mean and a limited variance in the absence of observations, the integrated EQ kernel will have a mean that depends mainly on the last location observed and a variance that grows over time. This is clearly preferable in tracking scenarios where the subject does not reliably return to a nominal position when not observed. As we seek a GP over positions rather than velocities, we integrate the EQ kernel w.r.t. each of its inputs \cite{smith2018gaussian},
\begin{align}\label{eq:DoubleIntegratedEQ}
K_{\int}(x,x')
&=\int_{0}^{x}\int_{0}^{x'} K(u,v)\,dv\,du \\
&=\; \int_{0}^{x}\int_{0}^{x'} s^{2}\exp\!\Big(-\frac{(u-v)^2}{2l^2}\Big)\,dv\,du\\
&= s^{2} \frac{l^{2}}{2}\Bigg[
g\!\Big(\frac{x}{l}\Big)
- g\!\Big(\frac{x-x'}{l}\Big)
+ g\!\Big(\frac{x'}{l}\Big)
- g(0)
\Bigg],
\end{align}
where $g(z) =z \, \sqrt{\pi} \, \operatorname{erf}(z) + e^{-z^2}$.

Note that the start index $0$ is arbitrary - we would recommend placing it sufficiently far back in time from the observations such that the model can explain the observed locations through integrating a small velocity in a way compatible with the kernel's hyperparameters (i.e. that doesn't require excessive speeds).

\subsection{Variational Inference}
We use variational inference \cite{blei2017variational} to seek a surrogate posterior $q(F)$ to approximate the unknown true posterior $p(F|Y)$ with variational parameters we know. Variational inference has the benefit of being computationally inexpensive when compared to methods such as Markov Chain Monte Carlo \cite{salimans2015markov} or Expectation Propagation \cite{minka2013expectation}, allowing for inference to be performed in the field upon retrieving the receiver.

The surrogate posterior $q(F)$ is optimized to most closely represent the true posterior by minimizing the Kullback-Liebler (KL) Divergence between them,
\begin{align}
    &\mathcal{D}_{KL}{\huge[}q(F)||p(F|Y){\huge]} = -\int q(F) \text{ log} \frac{p(F|Y)}{q(F)}dF \\
    &= -\int q(F) \text{ log } \frac{p(F,Y)}{q(F)p(Y)}dF \\
    &= -\int q(F) \text{ log } \frac{p(F,Y)}{q(F)}dF + \int q(F) \text{ log } p(Y) dF \\
    &= -\underbrace{\int q(F) \text{ log } \frac{p(F,Y)}{q(F)}dF}_{\text{ELBO}} + \text{log }p(Y)
\end{align}
The marginal likelihood $p(Y)$ is unknown, but can be regarded as constant w.r.t. the surrogate distribution. Therefore maximising the Evidence Lower Bound (ELBO) is equivalent to minimizing the KL divergence bounded by this constant \cite{blei2017variational}.

We use a vector of inducing variables $u$ \cite{quinonero2005unifying} which describe $F$, selected at inducing inputs $Z$ evenly spaced in time across the duration of the path. We still assume that $Y$ depends only on $F$. The posterior is therefore calculated as $p(F,u|Y) \propto p(Y|F)p(F|u)p(u)$. We assume that $u$ is able to sufficiently determine $F$ such that $p(F|u,Y) = p(F|u)$ \cite{uhrenholt2021probabilistic}, and using this assumption we should augment the ELBO to use the inducing variables,
\begin{align}
    &\mathcal{L} = \int\int q(F,u) \text{ log }\frac{p(F,u,Y)}{q(F,u)}dFdu \\
    & = \int\int q(F,u) \text{ log }\frac{p(Y|F)p(F|u)p(u)}{p(F|u)q(u)}dFdu \\
    & = \int\int q(F,u) \text{ log } p(Y|F)dFdu + \int q(u) \text{ log } \frac{p(u)}{q(u)} du \\
    & =\mathbb{E}_{q(F,u)}[\text{log }p(Y|F)] - \mathcal{D}_{KL}[q(u)||p(u)] \label{eq:ELBO}.
\end{align}

One can see that $\mathcal{L}$ now consists of a data fitting term and a prior term.

$\mathcal{L}$ is now iteratively maximised w.r.t. the variational parameters (mean $m$ and lower diagonal matrix $R$, to give the positive semi-definite covariance $RR^{\top}$) using doubly stochastic variational inference \cite{pmlr-v32-titsias14}: at each iteration multiple samples are drawn from $F_{i}\sim q(F)$ by sampling from $q(u)$ and computing $F_{i}$ from the resulting $u_{i}$. These are used to calculate the log likelihood of the observations $\text{log}\;p(Y|F_{i})$ and averaged to approximate the expectation of the first term of $\mathcal{L}$, which is,
\begin{align}
    \mathbb{E}_{q(F,u)}[\text{log }p(Y|F)] \approx \frac{1}{N}\sum^{N}_{i=1}\text{log }p(Y|F_{i}).
\end{align}

As the prior $p(u)$ and surrogate posterior $q(u)$ are both Gaussian distributed, their KL divergence in the second term of Equation \eqref{eq:ELBO} has a closed form solution.

JAX \cite{jax2018github}, an automatic differentiation framework, is then used to compute the gradient of $\mathcal{L}$ wrt to $m$ and $R$, and update them accordingly. When $\mathcal{L}$ reaches its global maximum, the surrogate distribution most closely matches the true posterior. The resulting surrogate distribution $q(F)$ can be used to compute the location distribution at any given time point.

\subsection{Likelihood}
The variational inference approach requires that we write an autodifferentiable function to compute $\text{log }p(Y|F_{i})$, the log likelihood of an observation, given a sample $F_{i}$.

AoA inference via the full probability distribution method (described in section \ref{chapter:fullprobmethod}) produces AoA observations consisting of a distribution over a finite set of angles. The rejection sampling method (section \ref{chapter:rejectionsampling}) returns an empirical set of samples of $\theta$ which can be used to return a single estimate of the AoA. The peak RSS finding method (section \ref{chapter:peakfinding}) only returns a single AoA. We present two likelihood functions to be used in the calculation of the ELBO, one for the full distribution and one for a single AoA.

\subsubsection*{AoA Probability Distribution Likelihood}
The probability distribution we compute in Section \ref{chapter:fullprobmethod} for the AoA $p(\theta|\bm{y},\bm{\gamma})$ is already nearly the likelihood function we need - we simply need to compute, for the sample $F_{i}$, what the angle to the receiver $\phi$ would have been and, in principle, simply report the value of $\text{log }p(\phi|\bm{y},\bm{\gamma})$.

We first note that, to find $\phi$ simply requires trigonometry: the point $\bm{p}$ on the sampled path $F_{i}$ at the time of this observation has the transmitter's location $\bm{l}$ subtracted from it. Then the angle relative to North is computed,
\begin{align}
    \phi = \text{arctan2}(\bm{p}_{North}-\bm{l}_{North},\bm{p}_{east}-\bm{l}_{east})
\end{align}

Our final step is to simply report $\text{log }p(\phi|\bm{y},\bm{\gamma})$ for this value of $\phi$. However we computed $p(\theta|\bm{y},\bm{\gamma})$ previously by evaluating $p(\bm{y}|\bm{\gamma},\theta)$ over a grid of $\theta\sim Uniform(0^\circ, 360^\circ)$ and normalizing. We need to now report $\text{log }p(\phi|\bm{y},\bm{\gamma})$ for an arbitrary $\phi$ such that it is still normalized, it is fast and it is possible to autodifferentiate.

To do this we linearly interpolate between the precomputed values of $\text{log }p(\theta|\bm{y},\bm{\gamma})$ e.g. if $\phi=33.2^{\circ}$, the result will be $\text{log }p(\theta=33|\bm{y},\bm{\gamma})\times0.8 \;+ \; \text{log }p(\theta=34|\bm{y},\bm{\gamma})\times0.2$. This allows the autodifferentiation tool in JAX to compute meaningful gradients. We assume that the precomputed values are close enough together that the gradient is well approximated by this interpolation.

\subsubsection*{Single AoA Likelihood}
In Sections \ref{chapter:rejectionsampling} and \ref{chapter:peakfinding} we described approaches for selecting a single angle $\theta$ rather than a distribution for each observation. For observations in this form (of a single AoA) we need an alternative likelihood function. One might just place a Gaussian over the angular difference between the observed angle and the proposed receiver location, this however penalises those predictions close to a transmitter where a small change in location will lead to a large change in angle.

We convert the single angle observation into a two dimensional line $L$ (using $\cos \theta $ and $\sin \theta$) described by a unit vector $\bm{v}$ extending from a transmitter's location $\bm{l}$. The likelihood of $L$ given a predicted receiver location $\bm{p}$ on a sampled path $F_{i}$ should depend on $\bm{p}$'s distance $d$, from $L$, as the further away a predicted location is from this line, the less likely we assume it to be. This is given as,
\begin{align}
   d = |\bm{v}\times(\bm{p}-\bm{l})| 
\end{align}

A Gaussian distribution $p(\theta|f) = N(d,0, \sigma^{2})$ is then used to compute the likelihood of observations. For this to be a valid likelihood $\int p(y|f)dy$ must equal a constant. This isn't quite true for this likelihood function, but can be shown to be approximately true for our problem (see Supplementary Material SII for detailed derivation).

\section{Experimental Design} \label{chapter:experimentaldesign}
To demonstrate the ability of our system to track the receiver using the methods discussed, we have designed several experiments described here, the results of which are presented in Section \ref{chapter:results}.

\subsection{Empirical Measurement of Transmitter ARP} \label{chapter:trainingdata}
For all experiments conducted, training data (used to empirically measure the transmitters' ARP for use in the full probability distribution and rejection sampling AoA inference methods) was collected from 10 minutes of continuous scanning of packets along with their RSS from a single rotating transmitter at The Ponderosa Park, Sheffield, UK. The receiver was positioned 30m due North from the transmitter with clear line-of-sight, and was powered by an external power supply to allow for high frequency (10ms scan interval) capturing of packets, ensuring that sufficient training data was collected. This data is then used when inferring AoAs from new RSS observations as described in Sections \ref{chapter:fullprobmethod} and \ref{chapter:rejectionsampling}.

All experiments described below are conducted in different locations to that which the training data was collected, ensuring we assess the ability of the system to generalise for unknown landscapes.

\subsection{AoA Inference Experiment} \label{chapter:AoAInferenceExperiment}
The first experiment aims to assess how accurately the three different AoA inference approaches can infer an AoA from a burst of $k$ received packets. It also investigates how varying $k$ affects the accuracy of AoA inference across the approaches, to explore to what extent $k$ can be minimized to save power. Data was collected at Norfolk Heritage Park, Sheffield, UK in which we placed the receiver at a known angle $\theta$ from North, 100m away from the transmitter with clear line-of-sight, and used the receiver to capture 100 bursts, 2000ms in length, with 200 packets per burst (the highest sampling rate possible on the DA14531). 

From this test data we randomly choose one of the bursts, subsample $k$ packets from the burst at uniformly spaced intervals, and apply the three approaches for AoA inference as described in section \ref{chapter:angleinference}. We calculate the angular difference between the inferred AoA and the known true value of $\theta$, and repeat 1000 times for all three AoA inference methods to calculate the MAE (mean absolute error) and standard deviation of the MAE in degrees for each. As the number $k$ packets per burst is a key consideration in the power consumption of the tag, we do this for several values of $k$ to present how the different approaches for AoA inference vary with the granularity of RSS measurements across a burst.

In these tests, the assumed Gaussian noise standard deviation $\sigma$ in the RSS values is chosen to be 6dB based on empirical observations. To allow us to also assess the accuracy of the full probability distribution method, we use the mode (the $\theta$ with the highest probability) to be the angle prediction we calculate the error for so that the methods are comparable - though this does discount some information present in the full probability distribution that may be beneficial to path reconstruction later. 

\subsection{Path Reconstruction Experiment} \label{chapter:pathreconexperiment}
The next experiment is conducted to assess the ability of the system to reconstruct the path of a receiver using a GNSS unit for ground truth, and compare how AoA observations inferred from our multiple AoA inference approaches and varying values of $k$ affect the resulting path reconstruction. This allows us to determine how few RSS measurements we can use to minimize power consumption while successfully reconstructing a path.

Data was collected at the Norfolk Heritage Park, Sheffield, UK - a park with complex topology differing by 10 meters between the highest and lowest point of the area we conducted our experiment in, some tree cover and buildings around the edges. This experiment consists of 5 complex paths and 1 straight line path, each around 3 minutes in length within a 200m × 200m area. The paths conducted were each in different directions and of different shapes. Four transmitters were set up $\sim$125m apart from each other. A researcher moved the receiver at a fast walking pace along the paths while also recording their movements with a mobile GNSS device. The tag was set to perform burst scans 2000ms in length with a scanning interval of 10ms, meaning up to 200 packets per 2 second burst could be received (the maximum sampling rate possible with the DA14531). We assume the bursts of samples are continuous i.e. every 2000ms there is a 2000ms burst.

The receiver was powered by an external power supply during data collection to allow for the dense scanning duty cycle of up to 200 packets per burst to be performed over the several minute span of each path. We can then, for each path, subsample $k$ packets from each burst at uniformly spaced intervals to evaluate performance between both the limited packets received at low values of $k$ that the supercapacitor can handle and higher values of $k$. By subsampling from the same underlying measurements we ensure that the exact same movement paths and attenuation generated the RSS measurements evaluated across different values of $k$

We then reconstruct a path using the variational inference approach discussed in section \ref{chapter:pathreconstruction}, from AoA predictions inferred using the AoA inference methods presented in Section \ref{chapter:angleinference}.

The doubly stochastic variational inference algorithm optimises 60 inducing points per axis, evenly spaced between the start and end time of the data collected for each path. The integral of the EQ kernel is used, with a length scale of 0.5 seconds and a scale factor of 0.5ms$^{-1}$, chosen to reflect smooth continuous movement and a velocity that is consistent with walking.

Once the ELBO converges, the paths are inferred at evenly spaced time points at which each of the GNSS data points were recorded, such that we can calculate point-for-point the error (in meters) between the GNSS path and the inferred path. These errors are used to calculate the MAE and the standard deviation of the MAE across all paths assessed for each value of $k$ and each AoA inference method.

\subsection{Duty Cycle Parameter Investigation} \label{chapter:dutycycleexperiment}
The limited capacity of a power source small enough to fit on a subject such as a flying insect necessitates careful use of the few scanning duty cycles available across the powered lifespan of the receiver. So far the experiments described have only explored the effect of changing the number of samples in each burst, but in reality more samples per burst means we must take fewer bursts (for the same energy budget).

This experiment tests if it is more beneficial to take less frequent, accurate AoA observations or a greater number of uncertain AoA observations. Being able to customise the trade-off between power usage and measurement accuracy is a design consideration also explored by other localization systems in which power is a key factor \cite{zhang2022demand}. We first define a synthetic path comprised of a sine wave shape (arbitrarily selected as a complex path to test) across a minute of time, and two transmitters positioned North and West of the path. We use a synthetic path here so that we know the true AoAs that the path generated and therefore can plot AoA inference accuracy against path reconstruction accuracy.

We generate $n$ AoA observations, inferred from a burst scan of $k$ RSS measurements, at linearly spaced intervals in time along the path. We do this by finding the angle from a transmitter at each time and taking $k$ samples from an ARP profiled at Weston Park, Sheffield, UK (collected in the same method as the training data was generated in Section \ref{chapter:trainingdata}) oriented to the matching angle. This ARP profile is only used to generate the RSS observations from our synthetic path; the training data for the AoA inference models is still the data that was collected at The Ponderosa, Sheffield as described in Section \ref{chapter:trainingdata}. We use a different ARP here to generate the synthetic path so that our synthetic RSS observations are not generated from the same training data used to infer an AoA, avoiding overfitting. The samples are taken equally spaced across a 2000ms duration, so as to simulate a 2000ms burst scan.

We simulate a budget of 100 RSS measurements (though the real receiver's budget is considerably larger) taken across the path. We then consider a number of values of $k$ such that our total set of observations varies between a few bursts containing many RSS measurements (resulting in confident AoA predictions) or many bursts containing very few RSS measurements (resulting in uncertain AoA predictions). For each value of $k$, we sample 100 sets of RSS measurements from our synthetic path, and for each set we perform AoA inference on each burst, reconstruct a path and calculate the total MAE of AoA inference and path reconstruction across the sets. This allows us to assess how trading off quality of AoA predictions for quantity of predictions affects our final path reconstruction.

\subsection{Application in Flying Insect Tracking} \label{chapter:beetracking}
To demonstrate the practical application of the system, we deploy it in a flying insect tracking field trial. The goal of the trial is to observe the ability of foraging \textit{Bombus terrestris} to find their way back to their nest \cite{goulson2001homing}.

An artificial \textit{Bombus terrestris} nest \cite{biobest} was placed in a copse of trees at Common Lane Open Space, Sheffield, UK. The nest was given around a week to begin foraging in the locale to allow workers to become familiar with the local environment, and following this on the 19th of August 2025, four transmitters were set up in the 250m × 250m area surrounding the nest site. This area contained a second copse of trees and highly complex topology with a difference of 16m of elevation between the highest and lowest points.

\begin{figure}[!t]
    \centering
    \includegraphics[width=2.5in]{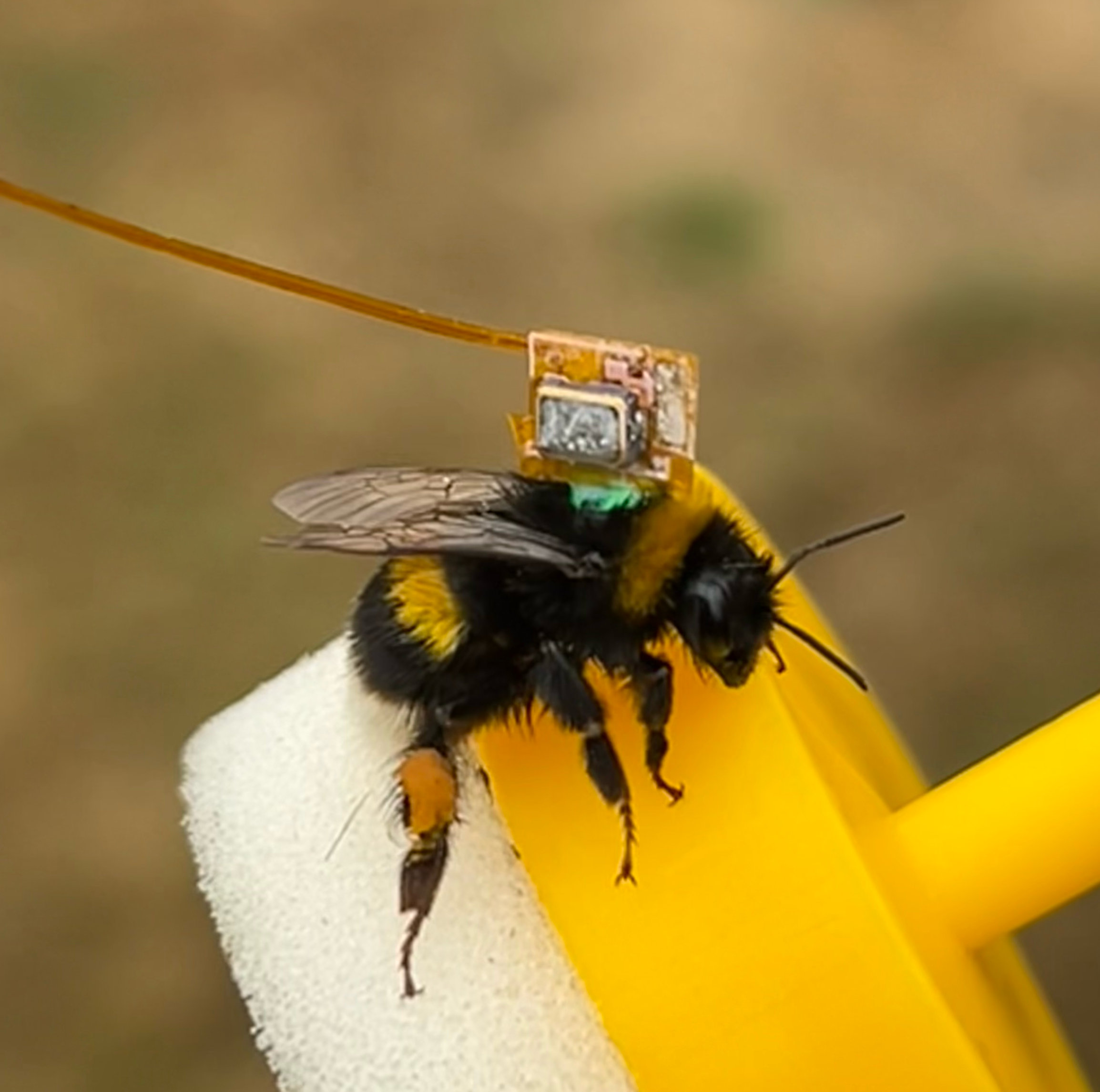}
    \caption{Image of \textit{Bombus terrestris} tagged with a receiver described in Section \ref{chapter:hardware} (glued to a green queen marking tag) sitting on the plunger of a queen marking pot following release during the experiment described in Section \ref{chapter:beetracking}.}
    \label{fig:BeeTag}
\end{figure}

A release site was designated $\sim$100m away from the nest. Foraging \textit{Bombus terrestris} returning to the nest were caught with a net and placed in a queen marking pot, then moved to the release site. The receiver was glued to a queen bee marking tag to increase the surface area with which the tag could contact the bee, and the tag was affixed to the caught bee's thorax using cyanoacrylate superglue (shown in Fig.~\ref{fig:BeeTag}).

The receiver was set to perform burst scans 1500ms in length consisting of $k = 10$ RSS measurements. The burst scans were performed 10 seconds apart from each other. From the retrieved data, we inferred AoA observations using the full probability distribution method, and used the integral EQ kernel with the same hyperparameter choices as those in Section \ref{chapter:PathReconstructionResults} in path reconstruction.

Finally the bee was released. Upon re-observation of a tagged bee at the nest the tag was removed, data was downloaded from the tag via BLE and the path the bee took from the release site to the nest was reconstructed.

\section{Results} \label{chapter:results}
\subsection{AoA Inference Accuracy}\label{chapter:AoAResults}
\begin{figure}[!t]
    \centering
    \includegraphics[width=3.25in]{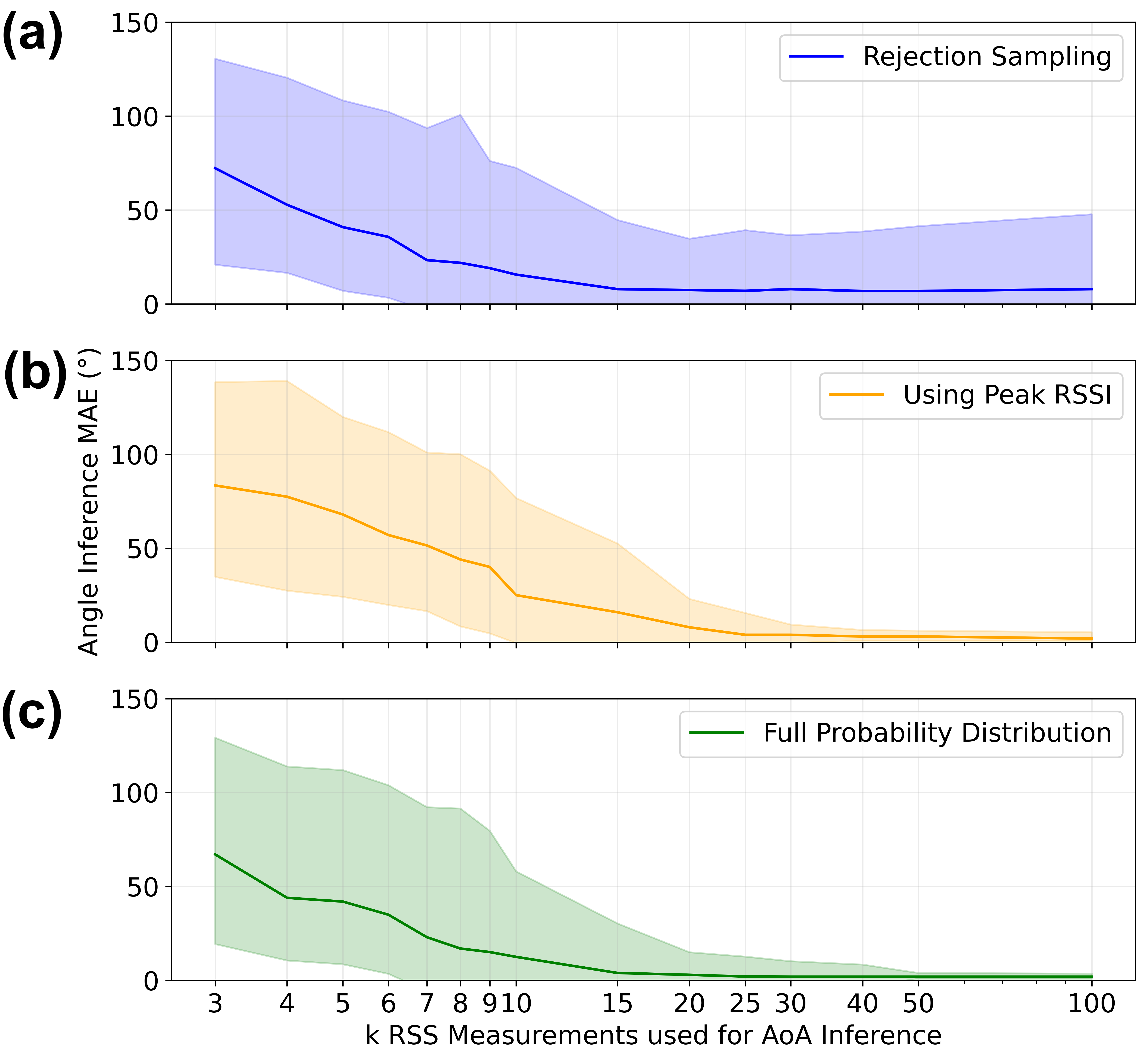}
    \caption{Plots showing the MAE in degrees of each AoA inference method across a range of $k$ number of RSS measurements used in the burst scan from which the angles were inferred. One standard deviation above and below the mean result is represented by the shaded areas. (a) MAE of rejection sampling method. (b) MAE of peak RSS method. (c) MAE of full probability distribution method. }
    \label{fig:AoAInferenceTestResults}
\end{figure}

We first assess the ability of the system to infer the AoA of a burst scan of packets received from a single transmitter as described in Section \ref{chapter:AoAInferenceExperiment}. The results are presented in Fig.~\ref{fig:AoAInferenceTestResults}.

These results show that all three methods result in a low mean angular error when a large number ($k$) of RSS measurements are taken in a burst. The full probability distribution and peak RSS methods produce an MAE of under 2$^\circ$ with a standard deviation of approximately 1$^\circ$. Rejection sampling is notably inconsistent at these high values of $k$, resulting in an MAE of 5$^\circ$ with a large standard deviation of around 40$^\circ$, where the other methods produce confidently correct predictions of $\theta$ as expected. This is because, at high values of $k$, there are very few training points which match all observed RSS values in the rejection sampling training set. While a simple solution could be collecting more training data, exponentially more data is required as $k$ increases and this would significantly increase the already high computational complexity of constructing the rejection record and sampling from it.

We are, however, aiming to use the minimum value of $k$ possible to save power on the tag, and all methods begin to notably decay in accuracy below 15 RSS measurements. The full probability distribution and rejection sampling approach produce an MAE of $\sim$50$^{\circ}$ at $k = 5$ packets, while the peak finding approach nears a random estimate of $\theta$, and all methods result in highly uncertain AoA predictions with a standard deviation of over $\sim$50$^{\circ}$.

This likely occurs as the limited number of samples $y$ taken at low values of $k$ are unlikely to fall in or close to the HPBW of the high-gain antenna - the $\sim$40$^{\circ}$ frontal lobe of the ARP constitutes only $11\%$ of the ARP and we scan over a 2000ms burst (one rotation of the 30 rpm transmitter), so as $k$ gets smaller than 10 the probability of missing the HPBW increases. Peak finding fails as the peak is largely absent from the data. Rejection sampling simply returns a high number of records constructed from regions of the ARP away from the main lobe. Both of these approaches at values of $k$ lower than 10 are too uncertain to be useful for later path reconstruction.

\begin{figure}[!t]
    \centering
    \includegraphics[width=2.5in]{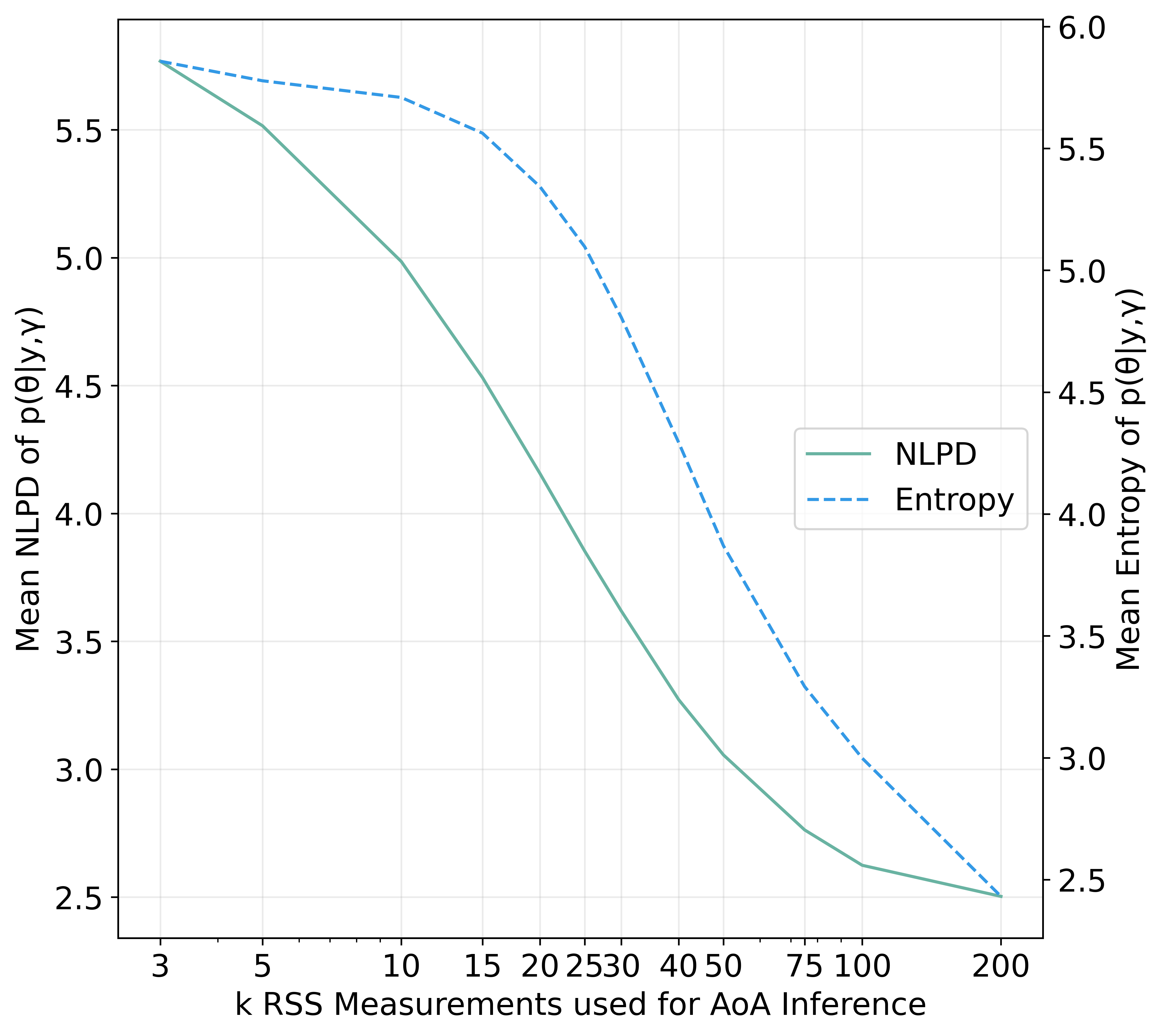}
    \caption{Plot of mean NLPD against mean entropy for our predicted angle distribution $p(\theta|\bm{y},\bm{\gamma})$ calculated for 1000 $\bm{y}$ subsampled from the test data across several values of $k$, showing that, as the predictive ability of $p(\theta|\bm{y},\bm{\gamma})$ decays with reducing values of $k$, its entropy increases therefore capturing the uncertainty of our reduced accuracy predictions.}
    \label{fig:NLPDvsEntropy}
\end{figure}

The full distribution method also suffers from highly uncertain AoA predictions at these low values of $k$, but has the added benefit of returning the full distribution of $p(\theta|\bm{y},\bm{\gamma})$. We perform another test to show that estimates of $p(\theta|\bm{y},\bm{\gamma})$ from the full probability distribution method that have a poor MAE due to using low values of $k$ are not confidently wrong and do capture the uncertainty of our prediction so that we can still use these uncertain estimates as input observations for path reconstruction. Using the method described in Section \ref{chapter:AoAInferenceExperiment}, we calculate the mean negative log predictive density (NLPD) of $p(\theta|\bm{y},\bm{\gamma})$ inferred from several subsampled sets of $\bm{y}$ across a number of values of $k$ as a measure of AoA inference effectiveness. We plot this against the mean entropy of these inferred $p(\theta|\bm{y},\bm{\gamma})$ distributions in Fig.~\ref{fig:NLPDvsEntropy} to show that estimates with poor predictive density have high entropy and therefore capture the uncertainty in predictions of $\theta$.

\subsection{Path Reconstruction Accuracy} \label{chapter:PathReconstructionResults}
We now assess the ability of the system to reconstruct the path of a receiver using a GNSS unit for ground truth through the experiment described in Section \ref{chapter:pathreconexperiment}. We have chosen not to assess the rejection sampling method (Section \ref{chapter:rejectionsampling}) further as it was shown in Section \ref{chapter:AoAResults} to have significantly poorer AoA inference accuracy than the other AoA inference methods (with substantially higher computational cost too).

An example of one of the paths reconstructed in this experiment and placement of the transmitters in the landscape is shown in Fig.~\ref{fig:examplepaths}, and all paths assessed are shown in Supplementary Material SIII.

\begin{figure}[!t]
    \centering
    \includegraphics[width=3.5in]{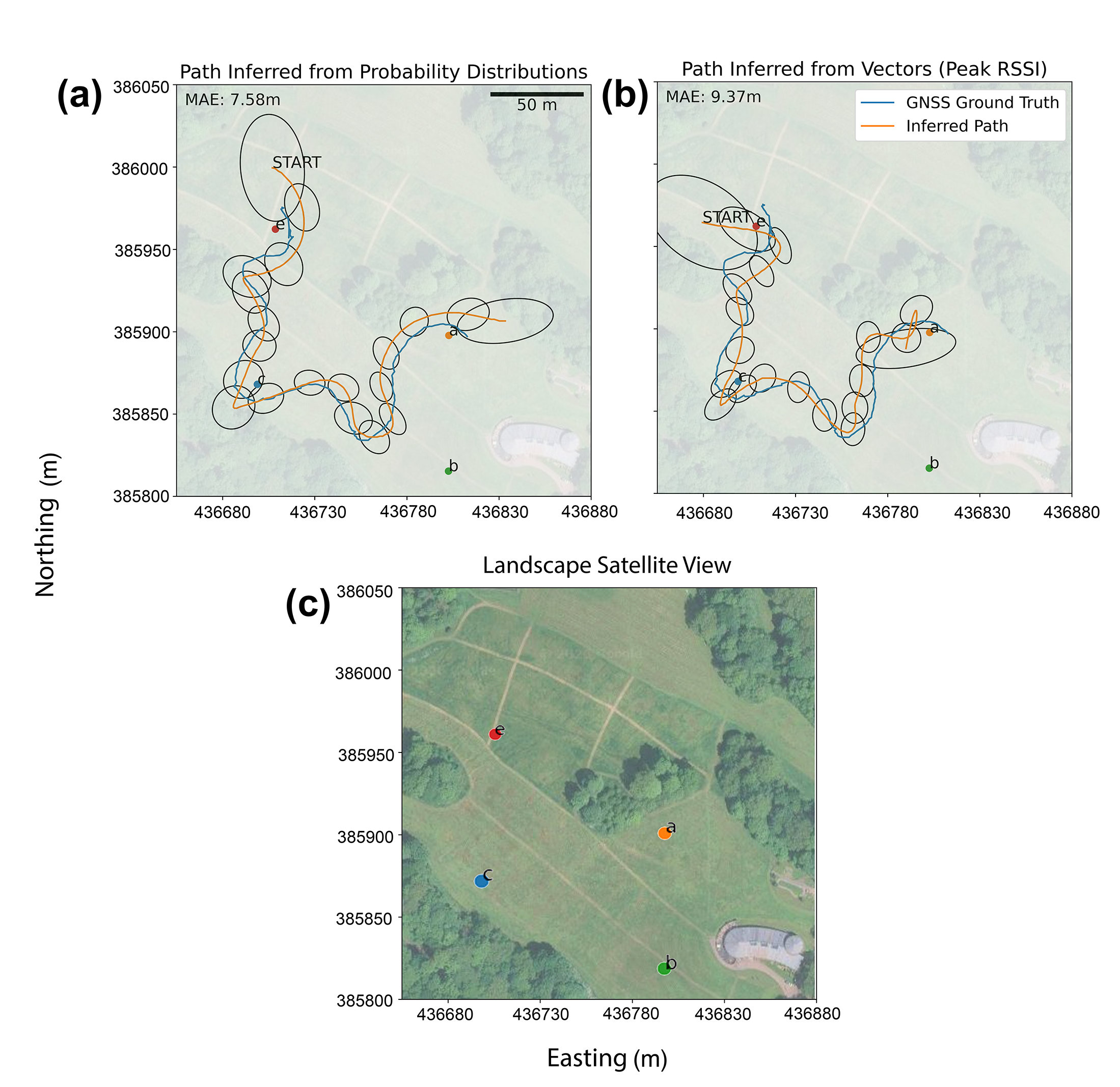}
    \caption{Example of one of the six paths used for assessing path reconstruction accuracy. The four transmitters are labelled, the blue path is the GNSS ground truth (3m accuracy), the orange path is the mean of the reconstructed path and the ellipses are 2 standard deviations of the covariance matrix at time points every 10 seconds along the inferred path. $k = 10$ samples were used per burst in these examples, and a path reconstructed from AoAs inferred via the (a) full probability distribution method and (b) peak finding method are presented. (c) A satellite view of Norfolk Heritage Park, Sheffield, UK at which the paths took place is also presented, retrieved from Google Maps \cite{googlemaps}. Note that the inferred paths' uncertainty grows at the beginning and end of the tracked paths, as before/after these points in time respectively there are no observations.}
    \label{fig:examplepaths}
\end{figure}

\begin{figure}[!t]
    \centering
    \includegraphics[width=3.5in]{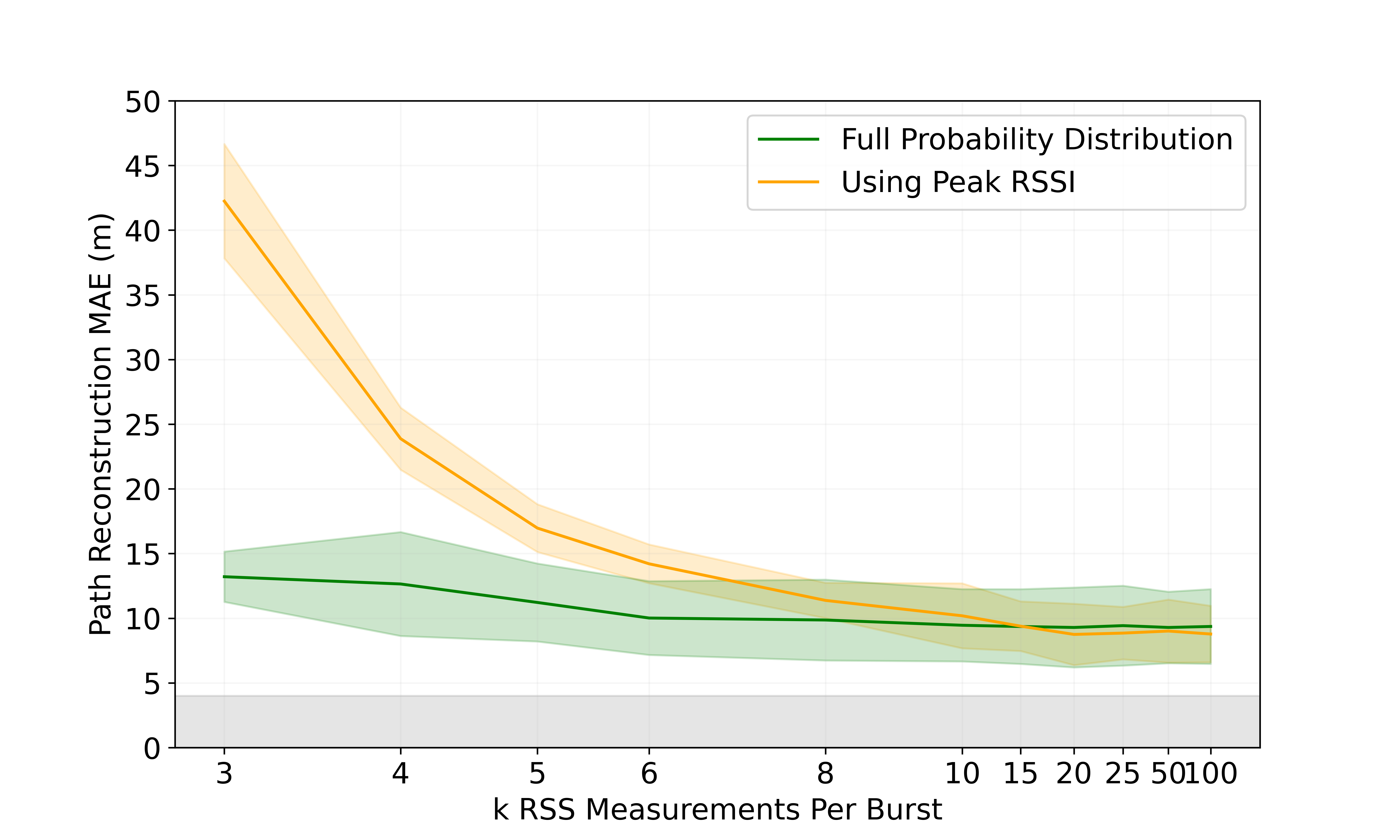}
    \caption{Plot of the MAE (m) of paths reconstructed using variational inference, from $k$ AoA observations inferred via two angle inference methods using parameter choices as presented in Section \ref{chapter:pathreconexperiment}. The coloured shaded areas represent one standard deviation above and below the MAE. The gray shaded indicates the estimated lower bound on the MAE achievable using this evaluation method. This bound arises from the positional uncertainty of the GNSS ($\approx$3m) used to collect the ground truthed path and the positions of the four transmitters. Assuming these errors are independent, their root-sum-of-squares yields a minimum error of $\sim$4m.}
    \label{fig:PathReconResults}
\end{figure}

Figure~\ref{fig:PathReconResults} presents the MAE computed by combining the six paths using either the full probability distribution method or peak finding method at a number of choices of $k$ to show how the degradation in AoA inference from taking very few RSS measurements observed in Section \ref{chapter:AoAResults} propagates to the accuracy of the inferred path.

Our results show that path inference can be successfully carried out to within 10m MAE using both assessed AoA inference methods when using a high enough number of packets $k$ per burst. However, we are aiming to use a minimal amount of power by using the smallest value of $k$ possible. The full probability distribution method is by far the most robust at low values of $k$, being able to reconstruct a path of movement within an accuracy of $\sim$15m in extremely power constrained scenarios where as few as 3 RSS samples per burst are taken.

Below 10 RSS measurements per burst, path reconstruction MAE from AoA observations inferred via peak finding begins to increase sharply as many bursts consisting of so few samples will not contain a measurement within the HPBW of the transmitters' antenna. This further demonstrates that using the full probability distribution of our AoA predictions at low $k$ values does contribute positively to reconstructing the receiver's path.

Across all values of $k$ the standard deviation of the MAE is approximately the same, suggesting that the error introduced by increasingly limited RSS sampling affects paths of different shapes equally.

\begin{figure}[!t]
    \centering
    \includegraphics[width=3.5in]{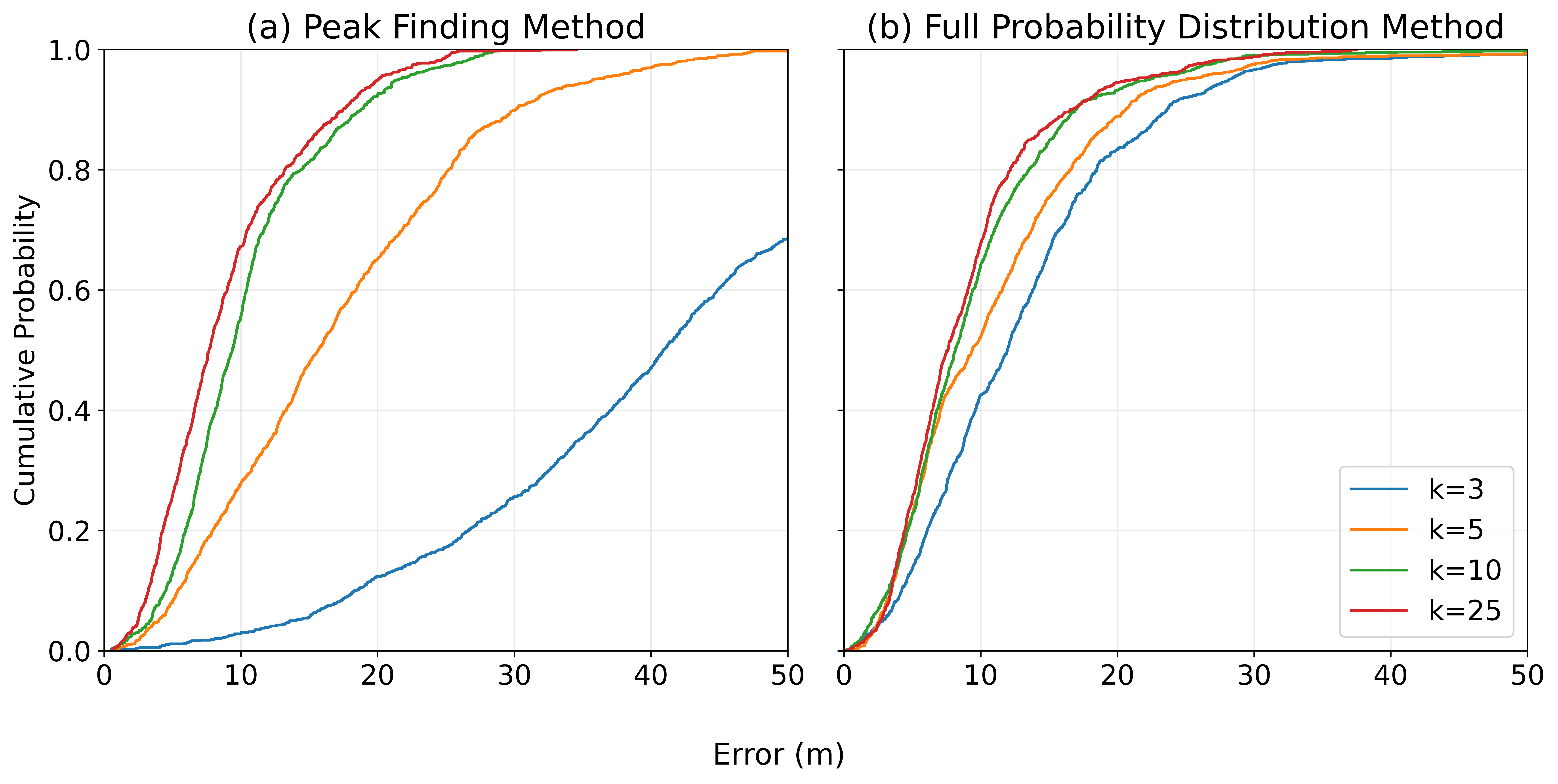}
    \caption{Empirical cumulative distribution plot of point-wise error in meters between all six reconstructed paths and their corresponding GNSS ground truth, from AoA observations inferred using (a) the peak finding method and (b) the full probability distribution method, at four selected values of $k$. The horizontal axis is truncated at 50m due to the poor performance of the peak finding method at $k=3$, approaching a random estimate of the ground truth path, and therefore having a large number of predictions exceed 50m of error.}
    \label{fig:CDF}
\end{figure}

To investigate the distribution of errors in the inferred paths, Fig.~\ref{fig:CDF} presents the cumulative probability distribution of the pointwise error in each inferred path at four selected values of $k$. 

Consistent with the results presented in Fig.~\ref{fig:PathReconResults}, both methods infer paths that have 80\% of their predictions within around 10m of error at high values of $k$. The full probability distribution method is largely insensitive to lower values of $k$, as at worst 80\% of predictions lie within around 16m of error when $k=3$. The peak finding method degrades much more severely as $k$ lowers. Both methods have a very small proportion of predictions that exceed 30m of error, explainable by the beginning and end of each reconstructed path where the absence of observations before or after the tracked period causes the posterior to revert towards the prior. The inferred paths such as the example presented in Fig.~\ref{fig:examplepaths} demonstrate that this is captured by a growth in uncertainty at the start and end of the reconstructed paths.

Based on the performance of the system at different values of $k$ we can also infer the lifespan of the 11mF supercapacitor powered receiver. The DA14531 chip used for the receiver draws 5mA while receiving, and has an operating voltage of between 1.8V and 3.3V in bypass power mode meaning that there is 16.5mC of usable charge. Since a 7ms duty cycle is used to sample every RSS measurement within a 2000ms burst, each RSS measurement contributes approximately 60$\mu$W to the average power consumption of the receiver in the case that we are constantly performing 2000ms bursts such as in this experiment. At the minimal $k=3$ samples per burst, with an accuracy of around 15m, the system will consume 180$\mu$W, powering the receiver for approximately 5 minutes of continuous tracking. At $k=10$ samples per burst, where the accuracy is around 10m, the system consumes 600$\mu$W and therefore the receiver lifespan reduces to approximately 90 seconds. This highlights that - while the power demand of the system is far lower than GNSS and existing RSS-based tracking methods which require many more samples - varying the duty cycle parameters to perform less frequent burst scans is critical to extend the powered lifespan of the tag.

\subsection{Trade-offs in Duty Cycle Parameters} \label{chapter:dutycycle}
\begin{figure}[!t]
    \centering
    \includegraphics[width=3.5in]{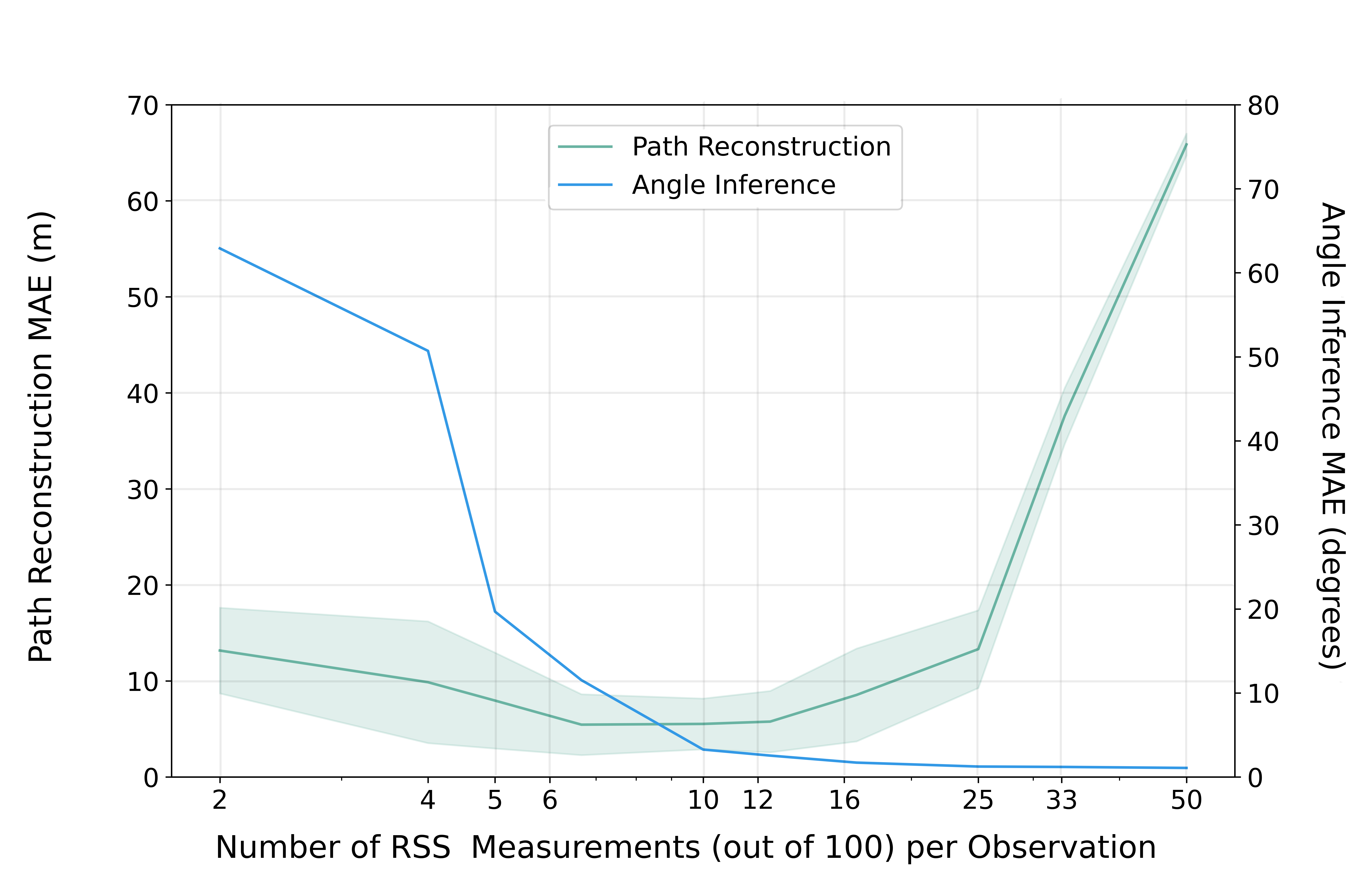}
    \caption{Plot of the MAE of AoA inference (blue) and resulting path reconstruction (green) from varying an allowance of 100 RSS measurements across values of $k$ (number of burst scans is 100/$k$) using synthetic data. The green shaded area is one standard deviation above and below the MAE of the path reconstruction.}
    \label{fig:dutycycleresults1}
\end{figure}

We investigate if it is beneficial to take less frequent, more accurate AoA observations (consisting of a higher $k$ number of packets per burst) or a greater number of uncertain AoA observations (lower $k$) - keeping the total energy used constant, using the experiment presented in Section \ref{chapter:dutycycleexperiment}. The results on a synthetic path, used for comparison of AoA inference accuracy and path reconstruction accuracy, are shown in Fig.~\ref{fig:dutycycleresults1}.

The results shows that the system is able to reconstruct a path accurate to within $\sim$10m, aligning with the results in Section \ref{chapter:PathReconstructionResults}, when given both numerous uncertain AoA observations and a smaller set of more accurate observations. This is important to note, as the powered lifespan of the receiver is limited, and the ability to control the duty cycle parameters is critical for customising the trade-off between the length of the tracked path and its reconstruction accuracy. It is only at the extremes of both cases that path reconstruction accuracy is degraded, though generally it appears to be better to have frequent uncertain AoA observations than too few AoA observations, shown by the sharp increase in error as the number of bursts lowers but the $k$ RSS measurements per burst increases, compared to the still relatively accurate $\sim$15m MAE at the inverse. This is likely because the extremely small number of AoAs inferred from high values of $k$ within the allowance cannot represent the path of movement that generated them.

\begin{figure}[!t]
    \centering
    \includegraphics[width=3.5in]{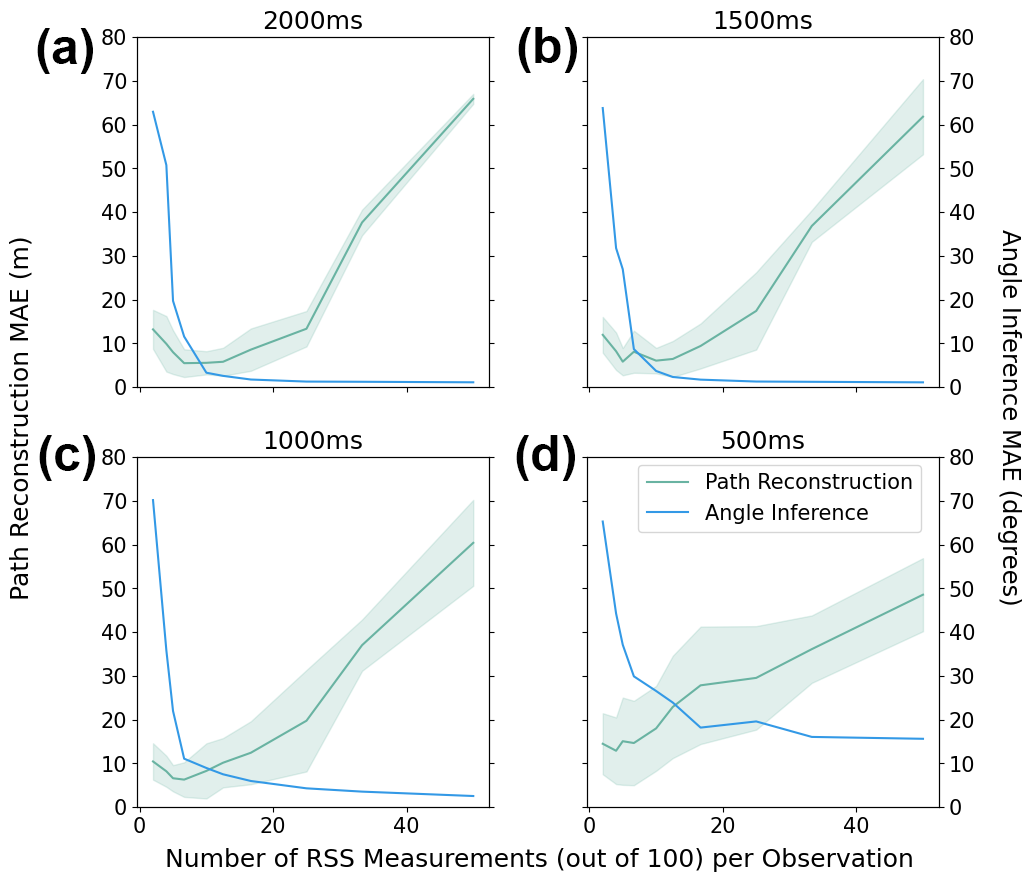}
    \caption{Plot of the MAE of AoA inference (blue) and resulting path reconstruction (green) from varying an allowance of 100 RSS measurements across values of $k$, for several burst scan lengths: (a) 2000ms, (b) 1500ms, (c) 1000ms, (d) 500ms. The green shaded area is one standard deviation above and below the MAE of the path reconstruction.}
    \label{fig:dutycycleresults2}
\end{figure}

A further variable of the duty cycle we can control is the length of the burst (the time over which our $k$ measurements are taken). We repeat the test with four selections of burst length in Fig.~\ref{fig:dutycycleresults2}, which shows that a 2000ms burst window results in the best path and angle reconstruction at low values of $k$. This is because the transmitters rotate every two seconds, so bursts substantially shorter than this will sometimes not include samples from the period when the highly informative HPBW portion of the transmitter's ARP is facing towards the receiver. These results are applicable in the case of our transmitter design which rotates at 30 rpm, though different transmitter rotation speeds would necessitate an adjusted burst length.

\subsection{Bombus Terrestris Flight Tracking Trial}
We demonstrate a major use-case of our system, tracking insect flight paths, by tagging bumblebee workers as they returned to their nest and displacing them in order to track their homing flight (see Section \ref{chapter:beetracking}).

\begin{figure}[!t]
    \centering
    \includegraphics[width=3.5in]{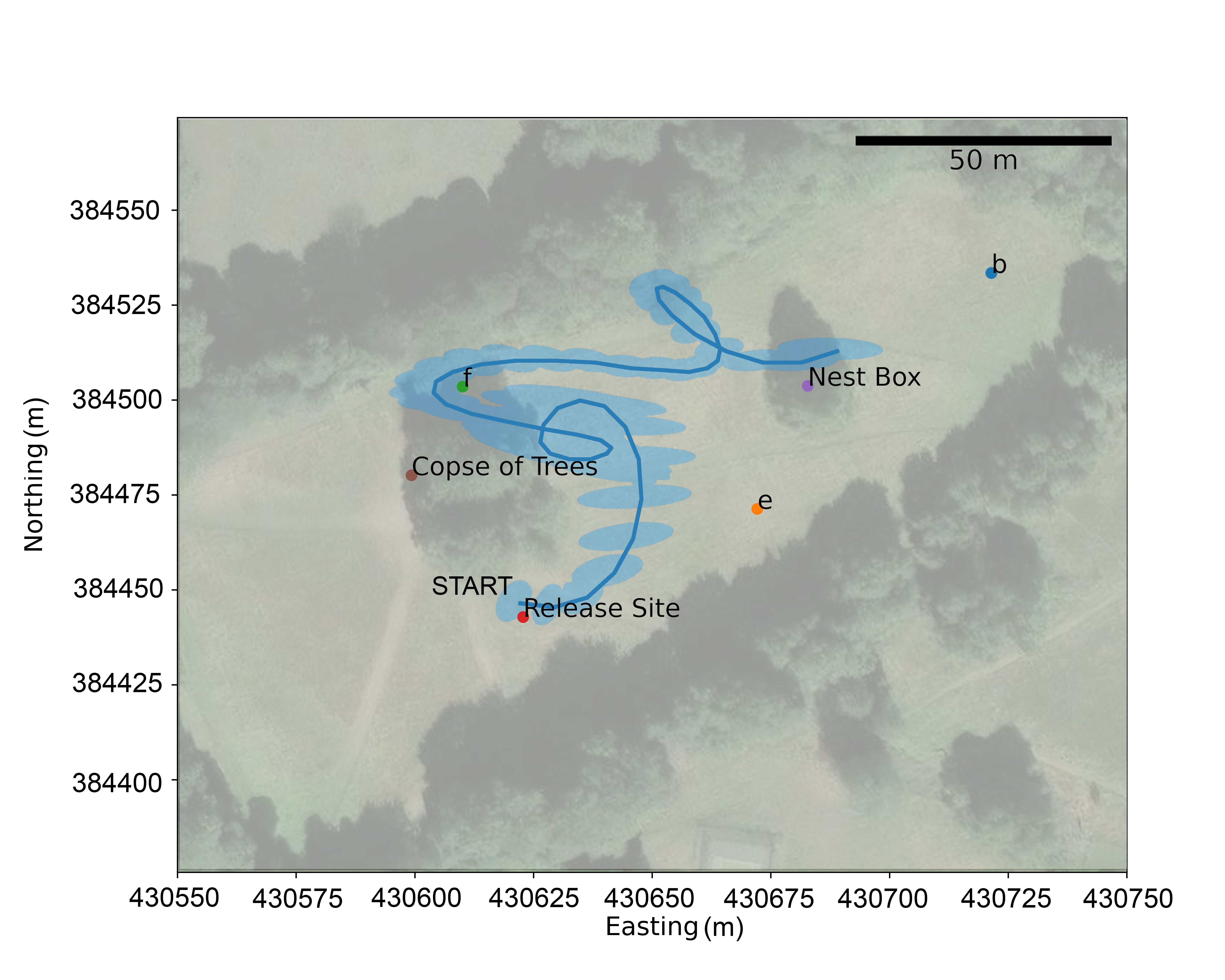}
    \caption{Reconstructed path (using full distribution AoA method and variational inference as described in Sections \ref{chapter:fullprobmethod} and \ref{chapter:pathreconstruction}) of a tagged \textit{Bombus terrestris} worker, caught at the Nest Box and displaced \& released at the Release Site. The bee after release was re-observed at the Nest Box and the path reconstructed from its retrieved tag successfully shows the bee flying from the Release Site to the Nest Box. The mean of the reconstructed path is represented by the blue line and the shaded blue region is the uncertainty encoded by the reconstructed path's covariance matrix ($^+_-$2 standard deviations). The three transmitters b,e and f are also shown, overlain onto a satellite image of Common Lane Open Space, Sheffield, UK retrieved from Google Maps \cite{googlemaps}.}
    \label{fig:beepath}
\end{figure}

Three tagged bees were re-observed within varying windows of time after release, ranging between a few minutes and multiple hours. Fig.~\ref{fig:beepath} presents one of the paths reconstructed from a bee re-observed around 10 minutes post release, showing a clear path the tagged bee took from the release site to the nest. Notably, researchers at the release site observed the tagged bee fly around the copse of trees immediately on release which matches the reconstructed path. The path also shows the tagged subject following the row of trees on the North side of the field, consistent with the well studied behaviour of \textit{Bombus terrestris} following linear landscape features \cite{brebner2021bumble}. Finally the reconstructed path shows the bee return to the nest as observed by fieldworkers when the tagged bee was retrieved at the nest site.

\section{Conclusion}
We have presented a novel RSS-based method for probabilistically reconstructing the movement path of an ultra low-power, lightweight receiver across a complex, unknown landscape using simple rotating high-gain transmitters. Through ground-truthed experiments we show $\sim$15m path reconstruction accuracy is achieved using $<$180$\mu$W with as few as 3 RSS measurements per AoA observation, improving to $\sim$10m at $<$600
$\mu$W with increased RSS measurements per observation.

This is achieved through two contributions, the first being a novel method of AoA inference by modelling RSS measurements from an anisotropic, rotating antenna with a Bayesian approach, allowing for probabilistic predictions of AoA to be made even with very few, noisy RSS measurements. The second is a probabilistic path reconstruction algorithm, able to reconstruct the receiver's movement path from uncertain, non-concurrent AoA predictions by modelling the path as a GP and performing inference using doubly stochastic variational inference \cite{pmlr-v32-titsias14}.

We demonstrate a use case of the system by tracking the return flights of displaced \textit{Bombus terrestris} foragers in a  complex environment. The ability to track bee movement and behaviour over a landscape scale is critical in advancing knowledge of behavioural ecology \cite{trappes2023tracking} and informing environmental conservation efforts \cite{chowdhury2023insects}, but previous tagged tracking methods \cite{riley1996tracking, fisher2021locating} have limitations in viability in complex landscapes and tag weight. Our 38mg receiver was light enough to be carried by the tagged forager, was sufficiently powered by its supercapacitor for long enough to capture the bee's full return flight, and the reconstructed path is consistent with field observations. The receiver is archival, meaning re-observation of the tracked subject is required to read off the received data, making the system particularly well suited to landscape-scale tracking of flying insect central place foragers in complex environments previously inaccessible to researchers.

\section*{Acknowledgments}
Special thanks to fieldworkers Euan Emery and Lauren Thomas.

\section*{Ethics Statement}
Under the UK Animals (Scientific Procedures) Act 1986, protected animals are defined as living vertebrates other than man, together with cephalopods. Invertebrates such as \textit{Bombus terrestris} used in this research fall outside this definition and their use in research does not require a Home Office licence or associated ethical review. The University of Sheffield's Animal Welfare and Ethical Review Body operates in compliance with this Act, and accordingly no institutional ethical review was required for this work. Notwithstanding this, we applied the principles of the three-Rs (Replacement, Reduction and Refinement) in our approach to experiments by using only one artificial nest box, handling individuals with care, removing tags from animals following recapture and noting that tagged bees were observed to return to flying and navigation after release.

\bibliographystyle{IEEEtran}
\bibliography{bibliography}
\setcounter{figure}{0}
\renewcommand{\thesection}{S\roman{section}}
\renewcommand{\theequation}{S\arabic{equation}}
\renewcommand{\thefigure}{S\arabic{figure}}
\setcounter{section}{0}
\setcounter{subsection}{0}

\section*{Supplementary Material}
\section{Marginalising Out an improper Random Variable Mean of a Gaussian that Lies on a Straight Infinite Line} \label{sup:marginalising}

Let $\bm{y}$ be a random variable, normally distributed with isotropic variance $\sigma^{2}$ and mean $\bm{x}$,
\begin{align}
    p(\bm{y}|\bm{x}) = N(\bm{y}|\bm{x},\sigma^{2}I).
\end{align}

We wish to marginalise out $\bm{x}$ which is a random variable defined using an improper prior such that it must lie upon a line that passes through the origin. To extend this to a line passing through any point, $\bm{p}$, simply repeat the derivation with a new RV equal to $\bm{y}-\bm{p}$.

\begin{figure}[h]
    \centering
    \includegraphics[width=3.5in]{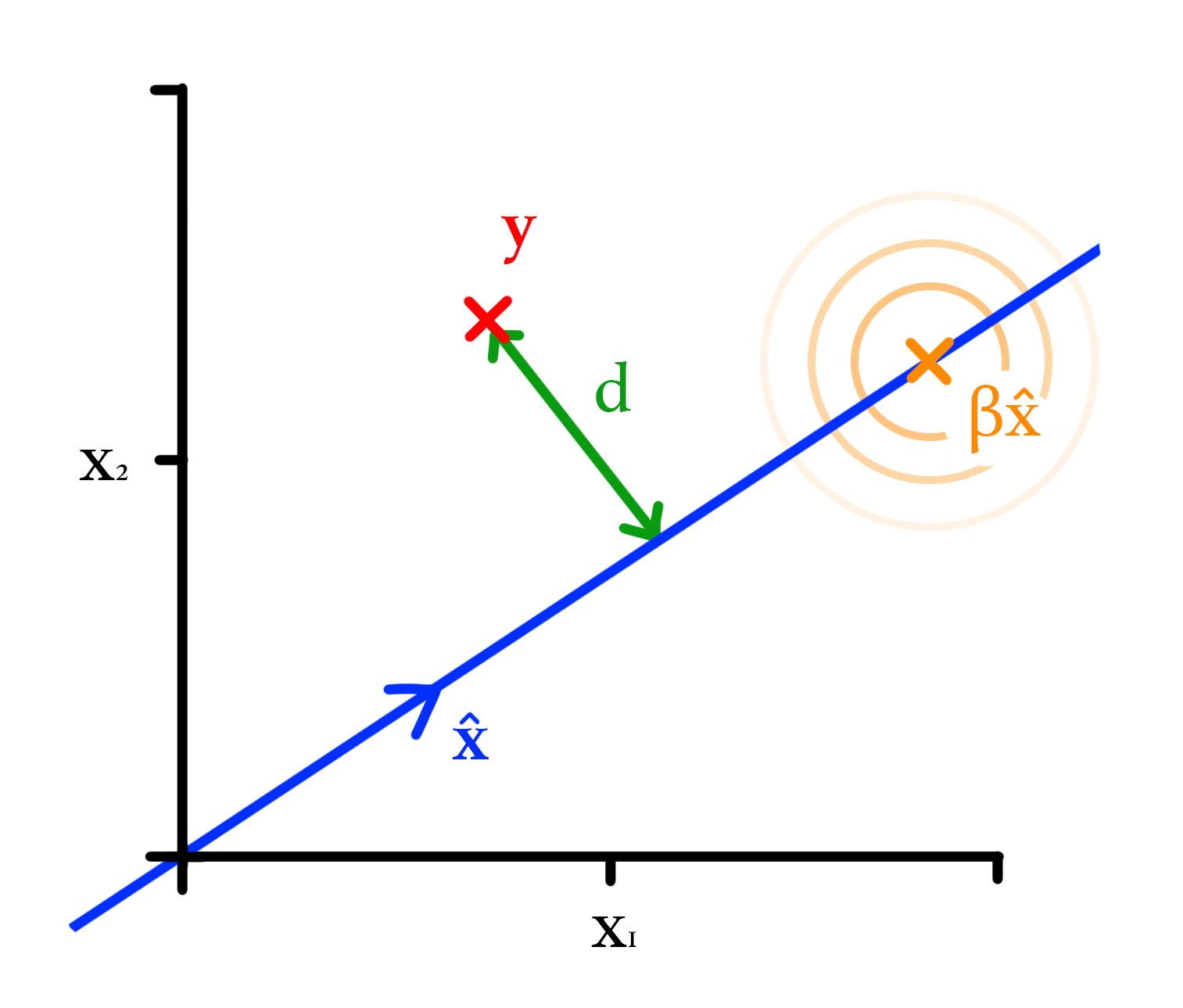}
    \caption{An illustration of the distance $d$ between the random variable $\bm{y}$ and the line defined by $\hat{\bm{x}}$.}
    \label{fig:aIntegrationProof1}
\end{figure}

So $\bm{x} = \beta\hat{\bm{x}}$ where $\hat{\bm{x}}$ is a unit vector and $\beta$ is a random variable $\beta\sim Uniform(-\infty, \infty)$. We define $d$ as the shortest distance between $\bm{y}$ and the line defined by $\beta\hat{\bm{x}}$, shown in Fig.~\ref{fig:aIntegrationProof1}.

We wish to show that the normalised marginal is,
\begin{align}
    p(\bm{y}) = N(d|0,\sigma^{2}).
\end{align}

This is equivalent to,
\begin{align}
    \int^{\infty}_{-\infty}p(\bm{y}|\beta\hat{\bm{x}},\sigma^{2}I)p(\beta)d\beta = N(d|0,\sigma^{2}).
\end{align}

\begin{figure}[h]
    \centering
    \includegraphics[width=3.5in]{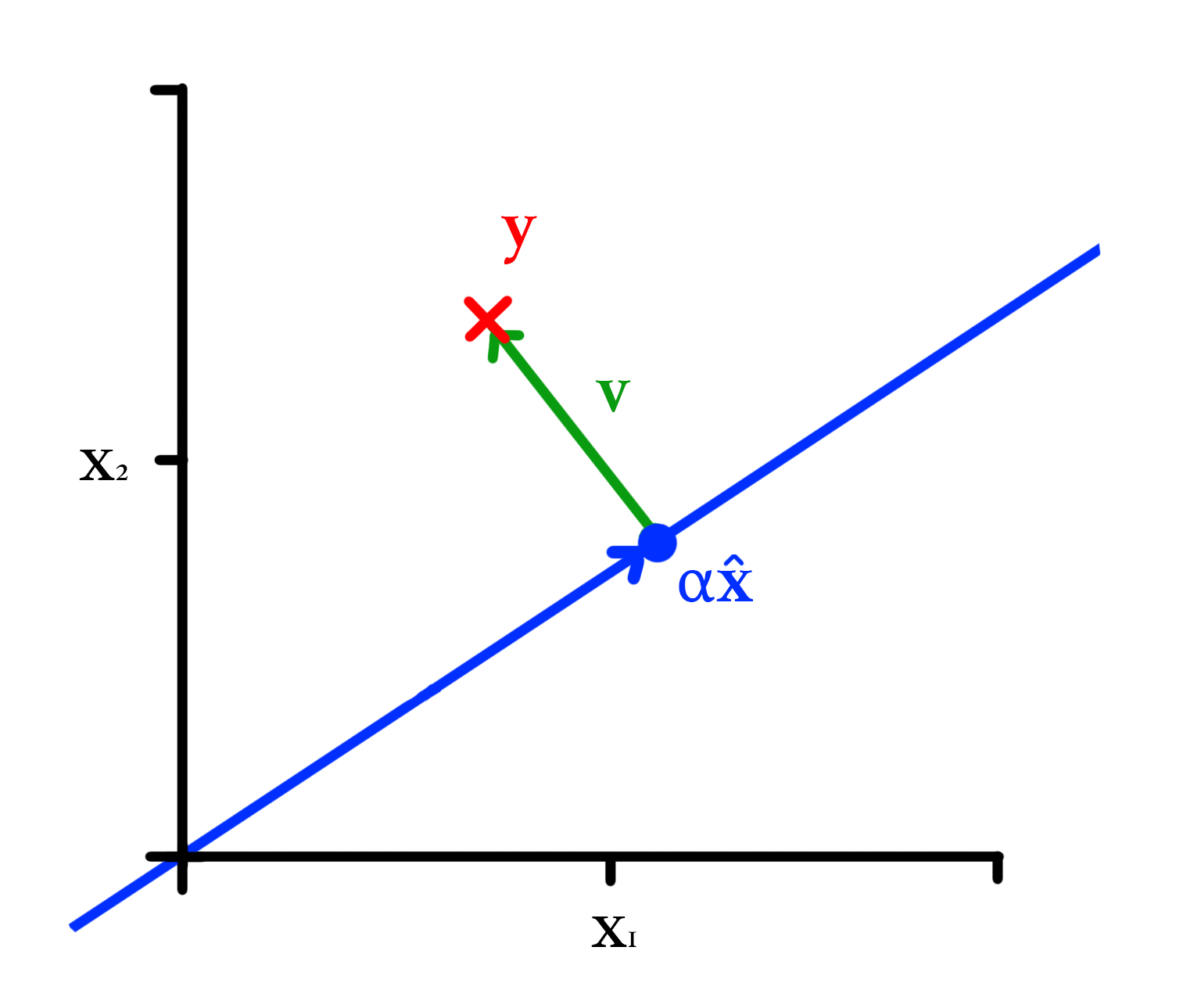}
    \caption{An illustration of the the point $\alpha\hat{\bm{x}}$ on the line defined by $\hat{\bm{x}}$ that is closest to $\bm{y}$.}
    \label{fig:aIntegrationProof2}
\end{figure}

We first write down $\bm{y}=\alpha\hat{\bm{x}} + \bm{v}$ where $\alpha\hat{\bm{x}}$ is the point on the line $\beta\hat{\bm{x}}$ that is closest to $\bm{y}$ as shown in Fig.~\ref{fig:aIntegrationProof2}. Therefore $\hat{\bm{x}}^{\top}\bm{v} = 0$ by definition.

Our probability density $p(\bm{y}|\bm{x})$ (which we will write $p(\bm{y}|\beta)$ to explicitly emphasise the dependence on the location along the line) is now,
\begin{align}
    p(\bm{y}|\beta)\propto\text{exp}\Large \left( \normalsize \frac{-(\alpha\hat{\bm{x}} +\bm{v}-\beta\hat{\bm{x}})^T(\alpha\hat{\bm{x}} +\bm{v} -\beta\hat{\bm{x}})}{2\sigma^{2}}\Large\right).
\end{align}

By also noting that $\hat{\bm{x}}^T\hat{\bm{x}} = 1$ and expanding out the product we obtain,
\begin{align}
    p(\bm{y}|\beta) \propto \frac{-(\alpha^{2} -2\alpha\beta+\beta^{2}+|\bm{v}|^{2}_{2})}{2\sigma^{2}}
\end{align}

So the probability density is,
\begin{align} \label{eq:aintegrationpd}
    p(\bm{y}|\beta)\propto\text{exp}\Large \left( \normalsize\frac{-(\alpha-\beta)^2}{2\sigma^2}\Large \right) \normalsize\text{  exp}\Large \left( \normalsize\frac{-|\bm{v}|^{2}_{2}}{2\sigma^2}\Large \right).
\end{align}

We can now marginalise (integrate out) $\beta$. We note that it has an improper (constant prior),
\begin{align}
    p(\bm{y}) \propto \int_{-\infty}^{\infty}e^{\frac{-(\alpha-\beta)^2}{2\sigma^2}}d\beta \times e^{\frac{-|\bm{v}|^{2}_{2}}{2\sigma^2}}.
\end{align}
We moved the second exponential of Equation \eqref{eq:aintegrationpd} outside the integral as it did not contain $\beta$.

This is invariate w.r.t. $\alpha$ as this is an integral over infinity (to see this, consider a change of variables, let $\epsilon = \beta - \alpha$, so $d\epsilon / d\beta = 1$, the integral becomes $\int_{-\infty}^{\infty}e^{-\epsilon^2/{2 \sigma^2}} d\epsilon = \text{const.}$). We note that $|\bm{v}|^{2}_{2}$ is $d^{2}$ so this is $\propto N(d|0,\sigma^{2})$.

\subsection{Applying this result to integrating out attenuation}\label{sup:attenuation}
Let $d$ be the distance between our vector of observed signal strengths $\bm{y}$ and a line $L$ in direction of $\bm{1}_{k}$ (a vector of ones, of length $k$) passing through the transmitter's predicted (unattenuated) signal strengths $\bm{t}$ for a given value of $\theta$. This line represents all possible expected signal strengths obtained by shifting $\bm{t}$ by the unknown attenuation $a$. $d$ measures how far observations $\bm{y}$ lie from the attenuated version of $\bm{t}$ that lies closest to $\bm{y}$.

To compute $d$, we define the difference vector $\bm{v} = \bm{t} - \bm{y}$ and remove its component along $L$. This is done by projecting $\bm{v}$ onto $L$ and subtracting it from $\bm{y}$, leaving a vector perpendicular to the line. The distance is therefore,
\begin{align} \label{eq:distance}
    d = \left\| \bm{v} - \frac{\bm{v}^\top \bm{1}_k}{k} \, \bm{1}_k \right\|.\end{align}
To summarise: We wish to integrate out $a$ from $p(\bm{y}|\bm{t}+a\bm{1}_k,\sigma^2)$. The shortest distance between the line defined by $\bm{t}+a\bm{1}_k$ is $d$. As shown in Supplementary Material SII this means that, if $a\sim Uniform(-\infty,\infty)$, then $p(\bm{y}|\bm{t}+a\bm{1}_k,\sigma^2) = p(d|0,\sigma^2)$.

Finally, to compute $d$ in a more convenient form we note that the second term of Equation \eqref{eq:distance} is the mean average of $\bm{v}$ as the product $\bm{v}^{\top}\bm{1}_{k}$ simply sums $\bm{v}$, and we divide by its length $k$. Substituting the definition of $\bm{v}$ in means that $d$ is simply,
\begin{align} \label{eq:thetalikelihood}
d = \Bigg |\Bigg|
\begin{bmatrix}
(t_{1}-\bar{\bm{t}}) - (y_{1}-\bar{\bm{y}}) \\
(t_{2}-\bar{\bm{t}}) - (y_{2}-\bar{\bm{y}}) \\
\vdots \\
(t_{k}-\bar{\bm{t}}) - (y_{k}-\bar{\bm{y}})
\end{bmatrix}
\Bigg |\Bigg|.
\end{align}
Hence, in Section IV-A, we define $d = ||(\bm{t}-\bar{\bm{t}})-(\bm{y}-\bar{\bm{y}})||_2$.

\section{Assumptions Needed to Ensure the Log Likelihood of Vector Observations is Approximately Valid} \label{sup:likelihood}
Ideally (with $\bm{f}$ fixed), $\int p(\bm{y}|\bm{f}), d\bm{y} = 1$. However, the integral over $\bm{y}$ is over a unit circle while the likelihood is a distribution over the length of the cross product with $\bm{f}$, so clearly it doesn't integrate to one. One could perform a change-of-variable operation although this is complicated by the awkward 1-to-2 variable mapping (from distance to vector) via the 2D cross product. Instead we show that the likelihood function does approximate a conditional distribution under some reasonable assumptions.

If one assumes w.l.o.g. that $\bm{f}$ lies along the y-axis, and the standard deviation ($\sigma$) of our distribution over the error distance' ($d$) is much less than the (predicted) distance of the receiver from the transmitter ($f_y$), then one can approximate the vector from the predicted location $\bm{f}-\bm{c}$ to the nearest point in the $\bm{y}$-direction as a 1-D scalar $a_x$ whose magnitude equals $d$. There is one angle to specify the direction of $\bm{y}$: $\sin \theta = a_x / f_y$. One can use the small-angle approximation to write that $f_y \theta \approx a_x$. It is required that $C \int_{-\infty}^{\infty} e^{-a_x^2/(2\sigma^2)} da_x = 1$ (for some normalising constant $C$). If one were to assume that this equality is approximately true over a small domain ($-r<a_x<r$), then one can immediately see that if $a_x$ were replaced with $f_y \theta$ one will still have a Gaussian, albeit with a different normalising constant (equal to $C f_y$). This would appear to be a problem (the conditional distribution of $y$ given $f$ should integrate to one). However, the computation of the ELBO uses the expectation of the \emph{log} likelihood, and so the normalising term becomes an additive constant. Problematically though this constant' is now proportional to $f_y$, which compromises the assumptions around the Monte Carlo approximation to the expectation. It is expected that the range of values for $f_y$ for a given computation of the expectation should be narrow (otherwise it would suggest there is no capability for estimating the receiver's location), so I propose that one can approximate that term with a single constant.

\section{Path Reconstruction Sample Results}\label{sup:pathexamples}
Presented in the following Figures in this section are the three 4 minute long paths that were used in assessing the path reconstruction algorithm in Section VI-C The GNSS paths have been presented alongside a path reconstructed from AoA observations produced from the peak finding and full probability distribution methods, via the variational inference method and parameters used in Section VI-C, across a selection of values of $k$ RSS measurements per burst.

\begin{figure*}[tp]
    \centering
    \includegraphics[width=5in]{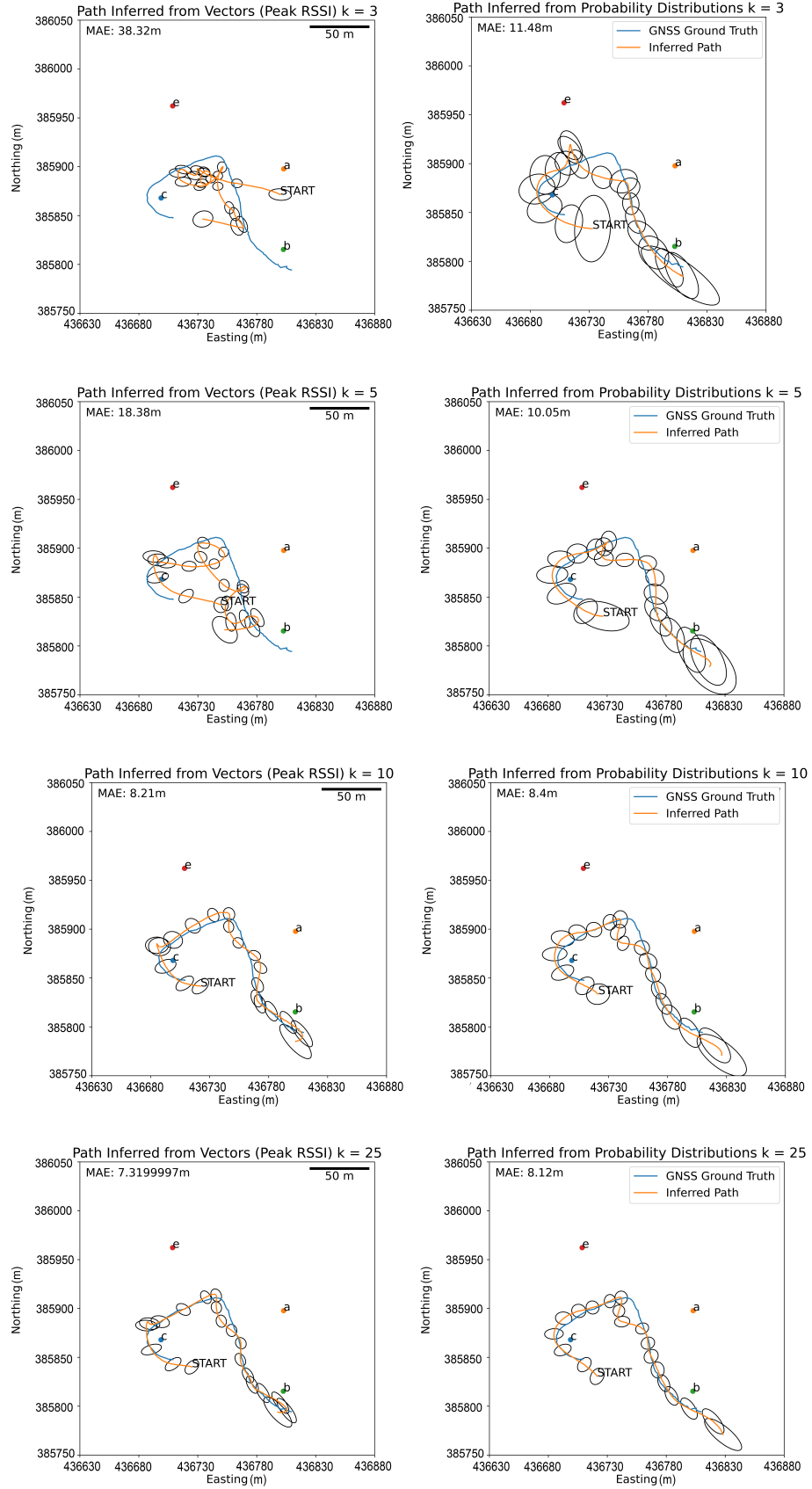}
    \caption{Plot of path one's GNSS ground truth (blue) and a reconstructed path's mean (orange) and standard deviation of its covariance matrix at 20 evenly spaced time points across the path (black ellipses). The transmitters a,b,c,e are plotted and labelled. Path reconstructions using $k = 3,5,10\text{ } \&\text{ }25$ packets per burst were used alongside the peak finding (left) and full probability distribution (right) AoA inference methods.}
    \label{fig:path1}
\end{figure*}

The first path is presented in Fig.~\ref{fig:path1}, the second path is presented in Fig.~\ref{fig:path2} and the third path is presented in Fig.~\ref{fig:path3}, the fourth path is presented in Fig.~\ref{fig:path4}, the fifth path is presented in Fig.~\ref{fig:path5} and the sixth path is presented in Fig.~\ref{fig:path6}, 

\begin{figure*}[tp]
    \centering
    \includegraphics[width=5in]{images/path2examples.png}
    \caption{Plot of path two's GNSS ground truth (blue) and a reconstructed path's mean (orange) and standard deviation of its covariance matrix at 20 evenly spaced time points across the path (black ellipses). The transmitters a,b,c,e are plotted and labelled. Path reconstructions using $k = 3,5,10\text{ } \&\text{ }25$ packets per burst were used alongside the peak finding (left) and full probability distribution (right) AoA inference methods.}
    \label{fig:path2}
\end{figure*}

\begin{figure*}[tp]
    \centering
    \includegraphics[width=5in]{images/path3examples.png}
    \caption{Plot of path three's GNSS ground truth (blue) and a reconstructed path's mean (orange) and standard deviation of its covariance matrix at 20 evenly spaced time points across the path (black ellipses). The transmitters a,b,c,e are plotted and labelled. Path reconstructions using $k = 3,5,10\text{ } \&\text{ }25$ packets per burst were used alongside the peak finding (left) and full probability distribution (right) AoA inference methods.}
    \label{fig:path3}
\end{figure*}

\begin{figure*}[tp] 
    \centering
    \includegraphics[width=5in]{images/paht4examples.png}
    \caption{Plot of path four's GNSS ground truth (blue) and a reconstructed path's mean (orange) and standard deviation of its covariance matrix at 20 evenly spaced time points across the path (black ellipses). The transmitters a,b,c,e are plotted and labelled. Path reconstructions using $k = 3,5,10\text{ } \&\text{ }25$ packets per burst were used alongside the peak finding (left) and full probability distribution (right) AoA inference methods.}
    \label{fig:path4}
\end{figure*}

\begin{figure*}[tp]
    \centering
    \includegraphics[width=5in]{images/path5examples.png}
    \caption{Plot of path five's GNSS ground truth (blue) and a reconstructed path's mean (orange) and standard deviation of its covariance matrix at 20 evenly spaced time points across the path (black ellipses). The transmitters a,b,c,e are plotted and labelled. Path reconstructions using $k = 3,5,10\text{ } \&\text{ }25$ packets per burst were used alongside the peak finding (left) and full probability distribution (right) AoA inference methods.}
    \label{fig:path5}
\end{figure*}

\begin{figure*}[tp]
    \centering
    \includegraphics[width=5in]{images/path6examples.png}
    \caption{Plot of path six's GNSS ground truth (blue) and a reconstructed path's mean (orange) and standard deviation of its covariance matrix at 20 evenly spaced time points across the path (black ellipses). The transmitters a,b,c,e are plotted and labelled. Path reconstructions using $k = 3,5,10\text{ } \&\text{ }25$ packets per burst were used alongside the peak finding (left) and full probability distribution (right) AoA inference methods.}
    \label{fig:path6}
\end{figure*}

\vfill

\end{document}